\documentclass[10pt,twocolumn,letterpaper]{article}

\usepackage[pagenumbers]{iccv} 
\usepackage{amssymb}
\usepackage{algorithm}
\usepackage{multirow}
\usepackage{algpseudocode}
\usepackage{pifont}
\usepackage{subfloat}
\usepackage{graphicx}
\usepackage[T1]{fontenc}
\newcommand{\cmark}{\ding{51}}%
\newcommand{\xmark}{\ding{55}}%
\definecolor{iccvblue}{rgb}{0.21,0.49,0.74}
\usepackage[pagebackref,breaklinks,colorlinks,allcolors=iccvblue]{hyperref}

\def\paperID{****} 
\def\confName{ICCV}
\def\confYear{2025}

\title{SG-LDM: Semantic-Guided LiDAR Generation via Latent-Aligned Diffusion}

\author{Zhengkang Xiang \quad Zizhao Li \quad Amir Khodabandeh \quad Kourosh Khoshelham \\
The University of Melbourne\\
}

\begin{document}
\twocolumn[
{
    \renewcommand\twocolumn[1][]{#1}%
    \centering
    \maketitle
    \vspace{-1.5em}
    
    \includegraphics[width=\linewidth]{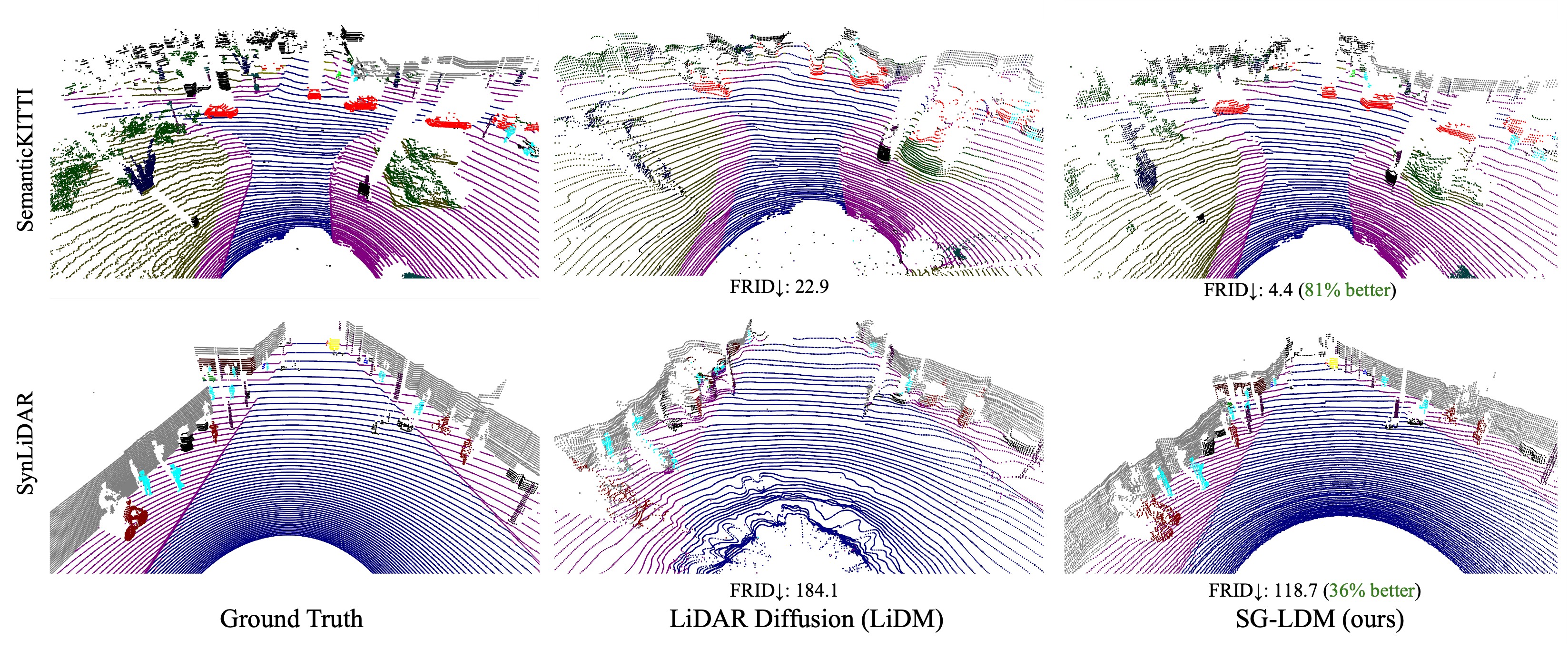}
    \vspace{-1.7em}
    
    \captionof{figure}{Quantitative comparison of our model and LiDM \cite{ran2024towards} on semantic-to-lidar conditional generation. Both models are trained on SemanticKITTI. Our approach achieves an improvement of 81\% in FRID score on the SemanticKITTI validation set and demonstrates robust generalization, yielding a 36\% FRID improvement on the SynLiDAR dataset.
    }
    \label{fig:teaser}
    \vspace{1em}
}
]

\begin{abstract}
Lidar point cloud synthesis based on generative models offers a promising solution to augment deep learning pipelines, particularly when real-world data is scarce or lacks diversity. By enabling flexible object manipulation, this synthesis approach can significantly enrich training datasets and enhance discriminative models. However, existing methods focus on unconditional lidar point cloud generation, overlooking their potential for real-world applications. In this paper, we propose SG-LDM, a Semantic-Guided Lidar Diffusion Model that employs latent alignment to enable robust semantic-to-lidar synthesis. By directly operating in the native lidar space and leveraging explicit semantic conditioning, SG-LDM achieves state-of-the-art performance in generating high-fidelity lidar point clouds guided by semantic labels. Moreover, we propose the first diffusion-based lidar translation framework based on SG-LDM, which enables cross-domain translation as a domain adaptation strategy to enhance downstream perception performance. Systematic experiments demonstrate that SG-LDM significantly outperforms existing lidar diffusion models and the proposed lidar translation framework further improves data augmentation performance in the downstream lidar segmentation task.
\end{abstract}    
\vspace{-1em}
\section{Introduction}

Lidar has clear advantages over RGB cameras in driving scene perception, including geometric accuracy and robustness to poor visibility and weather conditions. However, learning-based lidar perception systems require large-scale annotated datasets \cite{zhao2021point, wu2022point, lai2023spherical, kong2023rethinking, wu2024point,wu2024towards}, and manually assigning semantic labels or bounding boxes to each point (or cluster of points) is time-consuming and far more onerous than annotating 2D images. Additionally, existing real-world lidar datasets are naturally imbalanced, i.e., common classes like roads, buildings, and vehicles dominate the point clouds, while important but rarer classes (e.g. pedestrians, cyclists, traffic signs) are underrepresented \cite{behley2019semantickitti,caesar2020nuscenes}. This imbalance can bias learning algorithms, which tend to be overly tuned to frequent classes and struggle with minority classes. Due to these issues, models trained exclusively on real-world datasets suffer from biases and generalization gaps. 

To mitigate data scarcity and imbalance, leveraging synthetic data as a complementary resource has received considerable attention in recent years \cite{xiao2022transfer,xiang2024synthetic}. Advances in simulation platforms and generative modeling now enable the automated generation of large volumes of synthetic data, eliminating the need for labor-intensive data collection. Two primary approaches have emerged:
\begin{itemize}
\item \textbf{Physics-based simulation in virtual environments}: Leveraging high-fidelity simulators (e.g. CARLA \cite{dosovitskiy2017carla}, AirSim \cite{shah2018airsim}) built on game engines, this approach utilizes a virtual lidar sensor in a digital world to generate point clouds via ray-casting. The virtual environment can be populated with diverse 3D assets (roads, buildings, vehicles, pedestrians, vegetation, etc.), and since all objects are known, every point is automatically labeled. This approach eliminates the manual labeling effort while allowing complete control over the scene content.
\item \textbf{Data-driven generative models}: Beyond scripted simulation, deep generative models offer a novel way to synthesize lidar point clouds by learning the underlying data distribution. Recent works have explored using generative adversarial networks (GANs) \cite{caccia2019deep,nakashima2023generative} and diffusion models \cite{zyrianov2022learning, ran2024towards, hu2024rangeldm,nakashima2024lidar,wu2024text2lidar} to produce realistic 3D point patterns that mimic real lidar scans. These models can effectively generate synthetic lidar point clouds without requiring any pre-built 3D assets.
\end{itemize}
Despite their potential, both approaches face notable limitations. Virtual environments often exhibit a substantial domain gap compared to real-world data, necessitating additional training pipelines such as adversarial training \cite{tsai2018learning,vu2019advent,luo2019taking,luo2020unsupervised,wang2020classes,yuan2023prototype} or self-training \cite{saltori2022cosmix,yuan2024density} for domain adaptation. Existing lidar generative models \cite{caccia2019deep,zyrianov2022learning,hu2024rangeldm} focus on either unconditional data generation, which generates point clouds without incorporating semantic labels, or conditional point cloud upsampling and completion, where the process is guided by partial input data. However, both approaches lack the explicit semantic information necessary for effective data augmentation. Recently, LiDM \cite{ran2024towards} explored semantic-to-lidar synthesis, however, its performance remains underdeveloped. Specifically, the synthesized point clouds often display coarse geometric details, structural inconsistencies, and significant noise, which undermine their fidelity compared to real-world data. These limitations are clearly illustrated in \Cref{fig:teaser}.

We consider semantic-to-lidar a critical task, which has the potential to revolutionize 3D scene perception by enabling controllable, on-demand annotation of diverse scenarios. By conditioning lidar generative models on semantic segmentation maps, we directly specify the spatial arrangement of vehicles, pedestrians, and other scene elements, allowing for the synthesis of rich and complex lidar point clouds that align with predefined scene structures. Such controllable synthesis is particularly advantageous for addressing scenarios that are either rare or impractical for data collection in real-world driving, such as accidents, collisions, and ephemeral road incidents \cite{levering2020detecting}. This level of customizability and diversity would significantly enhance training datasets, improving model robustness and safety in autonomous systems. However, current methods for synthetic lidar data generation lack the ability to effectively utilize the rich semantic information available in both real and virtual environments.

To bridge this gap, we propose SG-LDM, a \textbf{S}emantic-\textbf{G}uided \textbf{L}idar \textbf{D}iffusion \textbf{M}odel. To effectively integrate classifier-free guidance, we propose a novel semantic alignment technique in the latent space that facilitates diffusion training in both unconditional and conditional modes. Prior diffusion‐based approaches \cite{ran2024towards, hu2024rangeldm} rely on latent diffusion architectures with a carefully crafted variational auto-encoder (VAE) and the diffusion model performs in the latent space. This incurs substantial compression loss and limited transferability, as these VAEs are typically trained with lidar data from a single source. In contrast, our approach discards the latent diffusion architecture, leading to significant performance improvements and better generalization. \Cref{fig:teaser} compares the lidar point clouds generated by a latent diffusion architecture (LiDM \cite{ran2024towards}) and our SG-LDM. LiDM exhibits noisy points around the ego vehicle due to suboptimal compression and fails to generalize across domains (trained on SemanticKITTI \cite{behley2019semantickitti} and tested on SynLiDAR \cite{xiao2022transfer}).

Moreover, we propose the first diffusion-based lidar translation framework built upon SG-LDM that leverages the inherent properties of the diffusion process to bridge the domain gap between real and synthetic lidar data. Unlike GAN-based translation frameworks \cite{xiao2022transfer}, our approach provides a more stable solution by effectively aligning both semantic and geometric features across domains. In summary, the contributions of this paper are as follows:
\begin{itemize}
    \item We present SG-LDM, a novel semantic-guided lidar diffusion model that establishes a new state-of-the-art in semantic-to-lidar generation.
    \item We propose a semantic alignment module in the latent space which improves performance of the diffusion model and enables effective classifier-free guidance.
    \item We introduce the first diffusion-based lidar translation framework built upon SG-LDM to bridge the domain gap between real and synthetic lidar data, offering a more stable alternative to GAN-based approaches.
    \item Through systematic evaluation of data generation and augmentation performance on the SemanticKITTI and SynLiDAR datasets, we demonstrate that SG-LDM significantly outperforms existing lidar generative models.
\end{itemize}



\section{Related work}
\textbf{Generative Modeling of 3D Point Clouds} Generative modeling of 3D point clouds has been an active research area for several years \cite{achlioptas2018learning, zamorski2020adversarial, yang2019pointflow, luo2021diffusion, zhou20213d, vahdat2022lion, wu2023fast}. More recent studies have proposed methods to condition point cloud generation on auxiliary modalities such as text \cite{nichol2022point} or images \cite{melas2023pc2}. Moreover, synthetic data produced by these generative models have been demonstrated to effectively augment downstream object recognition tasks \cite{xiang2024synthetic}.

For the generation of outdoor lidar point clouds, most methods require an initial transformation of the point clouds into a range map \cite{caccia2019deep, zyrianov2022learning, nakashima2024lidar, ran2024towards, hu2024rangeldm, wu2024text2lidar} or a bird's eye view representation \cite{xiong2023learning}. Although these representations typically contain geometry without semantics, LiDM \cite{ran2024towards} is the only method that addresses the semantic-to-lidar task using a latent diffusion framework.


\textbf{Diffusion Model} Diffusion models have become the dominant paradigm in computer vision generative tasks for both 2D and 3D datasets, underpinning many successful applications in image \cite{ho2020denoising,dhariwal2021diffusion,ho2022cascaded,rombach2022high,nichol2022glide,ramesh2022hierarchical,saharia2022photorealistic,peebles2023scalable}, video \cite{ho2022video,ho2022imagenvideohighdefinition,blattmann2023align,singer2023makeavideo,Wu_2023_ICCV,esser2023structure}, and 3D content generation \cite{luo2021diffusion,vahdat2022lion,jun2023shap,poole2023dreamfusion,lin2023magic3d,mo2023dit}. Since the introduction of the original denoising diffusion probabilistic model (DDPM) \cite{ho2020denoising}, many methods have been proposed to enhance the diffusion process. Among the most influential improvements are latent diffusion, denoised diffusion implicit models (DDIM), and classifier-free guidance (CFG). Latent diffusion \cite{rombach2022high} leverages a pre-trained variational autoencoder (VAE) to perform the diffusion process in a lower-dimensional latent space, thereby reducing computational complexity. DDIM \cite{song2021denoising} introduces a deterministic sampling procedure as opposed to the stochastic sampling in standard DDPMs and typically allows using significantly fewer inference steps (e.g., going from 1000 down to 50 or fewer) without completely sacrificing image quality. CFG \cite{ho2022classifier} strengthens the conditioning signal, enabling a controllable trade-off between fidelity and diversity in the generated outputs.

\textbf{Lidar Translation} Despite the extensive study of generative modeling-based image-to-image translation \cite{zhu2017unpaired,choi2018stargan,karras2019style,esser2021taming,gal2022stylegan,zhang2023adding,peng2023diffusion,acharya2023synthetic}, analogous techniques for lidar data remain largely underexplored. One strategy in lidar translation involves reconstructing a mesh from sequences of raw point clouds and subsequently employing ray casting to generate data in the target distribution \cite{lao2025lit}. However, mesh reconstruction introduces further domain discrepancies, primarily due to differences between the mesh representation and the intrinsic properties of raw lidar data. \citet{xiao2022transfer} proposed a GAN-based data translation method that uses two conditional GANs to translate the appearance and sparsity of lidar point clouds, respectively. \citet{yuan2024density} proposed a domain adaptive segmentation method by statistically transferring the density of the source point clouds to mimic the density distribution of the target point cloud. Several approaches explicitly model the lidar drop effect in real-world settings to enable synthetic-to-real domain adaptation \cite{nakashima2021learning, nakashima2023generative}. However, since these methods focus on a specific task, they cannot effectively address the general domain gap between the synthetic and real point clouds. 

\section{Method}
\begin{figure*}[!thbp]
\centering 
\includegraphics[trim=0.0cm 0.7cm 0.0cm 0.0cm, width=0.9\textwidth]{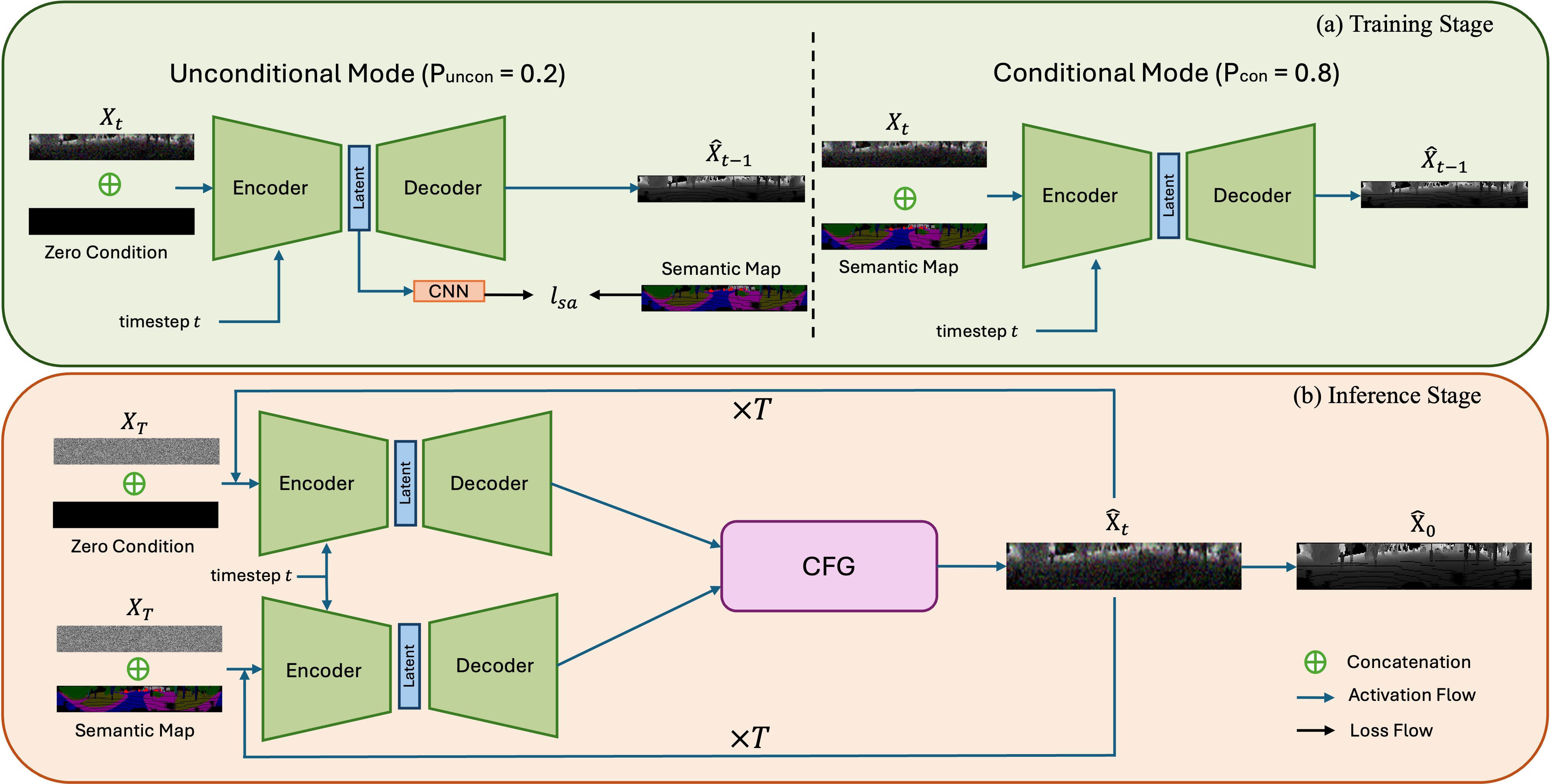}
\caption{Overview of SG-LDM. The model is trained in both conditional and unconditional modes, enabling classifier-free guidance during inference. In the unconditional mode, we incorporate a semantic alignment strategy in the latent space during training, which enhances the performance of unconditional generation and improves overall results.}   
\label{fig:overview}
\end{figure*}

In this section, we begin by outlining the problem formulation in \Cref{sec: PF}. Next, we introduce the core components of our diffusion model in \Cref{sec: DM}. We then introduce our semantic alignment module and detail the refined training process in \Cref{sec: SA}. Finally, we present a simple lidar translation framework based on our SG-LDM in \Cref{sec: DTF}. \Cref{fig:overview} presents the overview of the training and inference process of SG-LDM.

\subsection{Problem Formulation} \label{sec: PF}
\textbf{Semantic-to-Lidar Generation:} Given a labeled point cloud dataset $\mathcal{D} = \{X, Y\}$, where $X = \{ \mathbf{x}_i \in \mathbb{R}^3 \}_{i=1}^N$ is the set of points and $Y = \{\, \mathbf{y}_i \in \{1,\dots,K\}\}_{i=1}^N$ is the set of associated semantic labels, with each $\mathbf{y}_i$ taking one of $K$ possible class values, the task is to learn a generative model parameterized by $\theta$ such that the conditional density $p_{\theta}(X \mid Y)$ accurately captures the distribution of the points given the semantic labels.

\textbf{Lidar Data Representation}: Following existing work on lidar scene generative modeling \cite{caccia2019deep, zyrianov2022learning, nakashima2024lidar, ran2024towards, hu2024rangeldm}, we adopt the projected range image \cite{milioto2019rangenet++} as our lidar representation. The detailed range image and point cloud conversion is formulated in \Cref{sec: supp_range}. This approach relaxes the problem from 3D conditional generation to 2D conditional generation, enabling us to build our method on a mature 2D diffusion model. Rather than directly learning the conditional density of lidar point clouds $q(X \mid Y)$, we learn the conditional density of their 2D projections $q(\tilde{X} \mid \tilde{Y})$. For clarity, the random variables $X$ and $Y$ in the remainder of this paper refer to the projected range images and labels.

\textbf{Lidar Translation}: Given labeled lidar point cloud datasets from synthetic and real environments, denoted as $\mathcal{D}_s = \{X_s, Y_s\}$ and $\mathcal{D}_r = \{X_r, Y_r\}$, respectively, our goal is to translate synthetic point clouds \(X_s\) into \(\hat{X}_r\) so that they more closely resemble the real data $X_r$. Consequently, a model trained with the translated data \{\(\hat{X}_r\),$Y_s$\} for a downstream task like lidar segmentation is expected to achieve better performance in predicting $Y_r$ compared to a model trained solely on the raw synthetic data \{\(X_s\),$Y_s$\}.

\subsection{Revisiting Diffusion Model} \label{sec: DM}
We employ denoised diffusion probalistic model (DDPM) \cite{ho2020denoising} as the main training target and classifier-free guidance (CFG) \cite{ho2022classifier} during inference. In DDPM, a model is trained to reverse the noising steps of Markov chain, which is defined as the diffusion process by adding noise to clean data, also called forward diffusion process:
\begin{equation}\label{eq: Forward Diffusion}
    q(\mathbf{x}_t \mid \mathbf{x}_{t-1}) 
    = \mathcal{N}\!\bigl(\mathbf{x}_t;\, \sqrt{1-\beta_t}\,\mathbf{x}_{t-1},\, \beta_t \mathbf{I}\bigr)
\end{equation}
Here, $t\in\{1,\dots,T\}$ denotes the noising steps in the Markov chain, and the noise level at each step is governed by the parameters $\beta_{1},\dots,\beta_{T}$ generated according to the variance schedule. Given $\alpha_t = 1 - \beta_t$ and $\bar{\alpha}_t = \prod_{i=1}^t\alpha_i$, the original clean image $\mathbf{x}_0$ can be used to generate the noisy image at any time step $t$ via the following relation:
\begin{equation}
    q(\mathbf{x}_t \mid \mathbf{x}_0) =
\mathcal{N}\!\bigl(
  \mathbf{x}_t ;\, \sqrt{\bar{\alpha}_t}\,\mathbf{x}_0,\,
  (1 - \bar{\alpha}_t)\mathbf{I}
\bigr)
\end{equation}
A diffusion model parameterized by $\theta$ is trained to learn the reversed diffusion process:
\begin{equation} \label{eq: Backward Diffusion}
    p_{\theta}(\mathbf{x}_{t-1} \mid \mathbf{x}_t)
    = \mathcal{N}\!\bigl(\mathbf{x}_{t-1}; \mu_{\theta}(\mathbf{x}_t, t), \Sigma_{\theta}(\mathbf{x}_t, t)\bigr)
\end{equation}
Given $\epsilon \sim \mathcal{N}(0,I)$, the training objective of the diffusion model is derived by optimizing the variational lower bound, which leads to the following simplified loss function:
\begin{equation}
\mathcal{L}_{\text{DDPM}}(\theta)=\mathbb{E}_{\mathbf{x}, \boldsymbol{\epsilon}, t}\left\| \boldsymbol{\epsilon} - \boldsymbol{\epsilon}_{\theta} (\mathbf{x}_t + \boldsymbol{\epsilon}, t) \right\|^2
\end{equation}
Classifier-free guidance (CFG) \cite{ho2022classifier} is an essential technique in diffusion models that enhances conditional generation by amplifying the influence of conditioning information during the reverse diffusion process. Instead of relying solely on a conditional model $p_{\theta}(\mathbf{X} \mid \mathbf{Y})$, CFG also trains an unconditional model $p_{\theta}(\mathbf{X})$. During sampling, the noise predictions from both models are linearly combined:
\begin{equation}\label{eq: cfg}
    \tilde{\boldsymbol{\epsilon}}_{\theta}(\mathbf{x}_{t}, \mathbf{y}) = (1 + w) \boldsymbol{\epsilon}_{\theta}(\mathbf{x}_{t}, \mathbf{y}) - w \boldsymbol{\epsilon}_{\theta}(\mathbf{x}_{t})
\end{equation}
where we have $\text{guidance scale} = w+1$ . By increasing $w$, the reverse diffusion process is more strongly tied to condition $\mathbf{y}$, resulting in outputs with higher fidelity at the cost of reduced diversity. 

To train a diffusion model with both conditional and unconditional modes, a common practice is to set a probability for the unconditional mode, e.g. $\mathbf{P}_{uncon}=0.2$, which means that the training process will have a 20\% chance of having the semantic condition $\mathbf{y}$ replaced by a null condition $\varnothing$.
\subsection{Semantic Alignment} \label{sec: SA}
During experiments with CFG on a standard diffusion model, we observed a marked decline in performance after incorporating CFG. A closer analysis revealed that the model’s ability to generate content unconditionally was severely impaired when trained alongside its conditional counterpart. We hypothesize that the model was inadvertently optimized to leverage conditioning cues, so that when these cues are absent, it lacks sufficient information to produce coherent outputs. To address this, we introduce a semantic alignment module for the unconditional mode, ensuring that it can extract and utilize semantic information from raw inputs even without the explicit condition.


We incorporate a three-layer convolutional neural network $h_{\phi}$ that operates on the latent features produced by the diffusion encoder $f_{\theta}$. This network projects these features into a space that matches the dimensions of a downscaled semantic map. We then enforce semantic consistency by applying a cosine similarity loss between the projected features and the semantic map. This alignment loss guides the model to preserve robust semantic representations, even when explicit conditioning information is absent:
\begin{equation}
    \mathcal{L}_{\text{SA}}(\theta, \phi) = - \mathbb{E}_{\mathbf{x}, \boldsymbol{\epsilon}, t} 
\left[ \cos(\mathbf{y}_{i}, h_{\phi}(f_{\theta}(\mathbf{x}_t)) \right]
\end{equation}  
Additionally, to ensure that the semantic alignment is only applied when it is meaningful, we modulate its influence dynamically. Since it is uninformative for the encoder to extract features from predominantly noisy inputs at large timesteps, we introduce a dynamic weight $\lambda=1-\frac{\mathbf{t}}{\mathbf{T}}$. This schedule gradually decreases the alignment strength as the noise level increases. The final training objective for our SG-LDM in the unconditional mode becomes:
\begin{equation}
    \mathcal{L} = \mathcal{L}_{\text{DDPM}} + \lambda\mathcal{L}_{\text{SA}}
\end{equation}
Combining DDPM and semantic alignment, the training algorithm of our SG-LDM becomes:
\begin{algorithm}[H]
\caption{DDPM Training with Semantic Alignment}
\label{alg:training_dual_loss_if_else}
\begin{algorithmic}[1]
    \State \textbf{repeat}
    \State Sample \((x_0, y) \sim q(x_0,y)\)
    \State Sample \(t \sim \mathrm{Uniform}(\{1,\dots,T\})\)
    \State Sample \(\epsilon \sim \mathcal{N}(0,I)\)
    \State Compute noisy sample:
    \Statex \(\quad x_t = \sqrt{\bar{\alpha}_t}\,x_0 + \sqrt{1-\bar{\alpha}_t}\,\epsilon\)
    \State Sample \(u \sim \mathrm{Uniform}(0,1)\)
    \If{\(u < 0.8\)}
        \State \(\mathcal{L} = \bigl\|\epsilon - \epsilon_\theta(x_t,t,y)\bigr\|^2\)
        \State Take gradient step on \(\nabla_\theta \mathcal{L}\)
    \Else
        \State \(\mathcal{L}_{\text{DDPM}} = \bigl\|\epsilon - \epsilon_\theta(x_t,t,\varnothing)\bigr\|^2 \)
        \State \(\mathcal{L}_{\text{SA}}=-\mathrm{cos}(\mathbf{y}_{i}, h_{\phi}(f_{\theta}(\mathbf{x}_t)))\)
        \State \(\lambda=1-\frac{\mathbf{t}}{\mathbf{T}}\)
        \State \(\mathcal{L} = \mathcal{L}_{\text{DDPM}}+\lambda\mathcal{L}_{\text{SA}}\)
        \State Take gradient step on \(\nabla_{\theta,\phi} \mathcal{L}\)
    \EndIf
    \State \textbf{until} converged
\end{algorithmic}
\end{algorithm}

For the sampling process, we follow the standard classifier-free guidance using the linear combination (\cref{eq: cfg}) as illustrated in \cref{fig:overview}. This approach combines conditional and unconditional noise predictions, effectively improving the fidelity of the generated lidar samples conditioning on semantic maps.

\subsection{Lidar Translation Framework}\label{sec: DTF}
\begin{figure*}[!thbp]
\centering 
\includegraphics[width=0.8\linewidth]{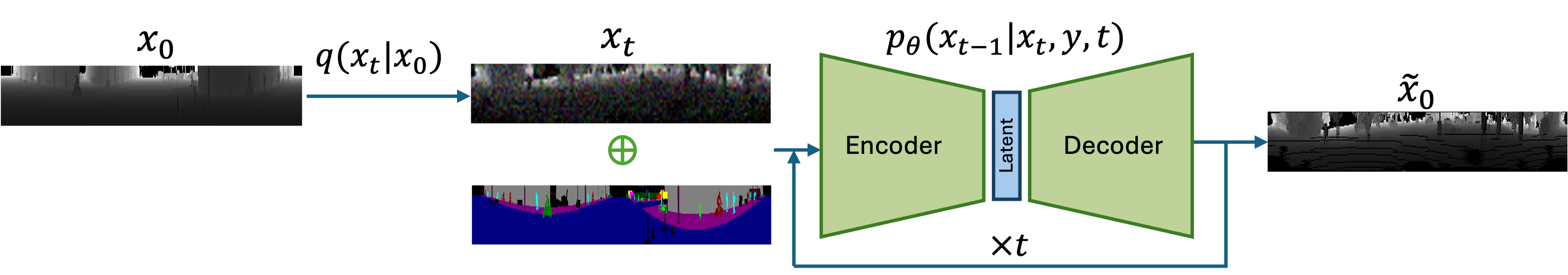}
\caption{Lidar Translation Framework. Forward diffusion is applied to the source data $x_0$ to progressively degrade its information, while backward diffusion reconstructs the target data $\tilde{x}_0$ from an intermediate noisy state $x_t$. The timestep $t$ represents the noise level, indicating the number of diffusion steps executed in the process. }   
\label{fig:method_LTF}
\end{figure*}

The diffusion model naturally bridges different data domains by progressively destroying the original information until it converges to a Gaussian distribution. Building on this concept, we propose a novel lidar translation framework based on SG-LDM. This framework is visually demonstrated in \Cref{fig:method_LTF}. Specifically, we apply the forward diffusion process (\cref{eq: Forward Diffusion}) to degrade information in the synthetic dataset and then use the reverse diffusion process (\cref{eq: Backward Diffusion}) to reconstruct lidar point clouds that conform to the target distribution from the intermediate states. This transformation can only be achieved with a semantic-to-lidar generative model. Unconditional models fail to retain the original semantic information, leading to the loss of semantic labels. By incorporating semantic labels as constraints, our approach ensures that the reverse diffusion process does not generate extraneous objects, thereby preserving the per-point labels essential for effective data augmentation.

\section{Experiments}
\begin{table*}[ht]
    \centering
    \resizebox{0.9\linewidth}{!}{
    \large
    \begin{tabular}{l|ccccc|ccccc}
        \toprule
        \textbf{Method} 
        & \multicolumn{5}{c|}{\textbf{SemanticKITTI \cite{behley2019semantickitti} (Same Domain)}} 
        & \multicolumn{5}{c}{\textbf{SynLiDAR \cite{xiao2022transfer} (Different Domain)}} \\
        \cmidrule(lr){2-6} \cmidrule(lr){7-11}
        & \textbf{FRID} $\downarrow$ 
        & \textbf{FSVD} $\downarrow$ 
        & \textbf{FPVD} $\downarrow$ 
        & \textbf{JSD} $\downarrow$ 
        & \textbf{MMD} ($\times10^{-4}$) $\downarrow$
        & \textbf{FRID} $\downarrow$ 
        & \textbf{FSVD} $\downarrow$ 
        & \textbf{FPVD} $\downarrow$ 
        & \textbf{JSD} $\downarrow$ 
        & \textbf{MMD} ($\times10^{-4}$) $\downarrow$ \\
        \midrule
        LiDARGen \cite{zyrianov2022learning} & 42.5 & 31.7 & 30.1 & 0.130 & 5.18 & - & - & - & - & - \\
        Latent Diffusion \cite{rombach2022high} & 24.0 & 21.3 & 20.3 & 0.088 & 3.73 & - & - & - & - & - \\
        LiDM \cite{ran2024towards} & 22.9 & 20.2 &17.7 & \textbf{0.072} & 3.16 & 184.1 & 147.9 & 144.3 & 0.277 & \textbf{1.48} \\
        \midrule
        SG-LDM (ours) & \textbf{4.4} & \textbf{10.5} & \textbf{7.9} & 0.084 & \textbf{1.31} & \textbf{118.7} & \textbf{78.0} & \textbf{80.3} & \textbf{0.145} & 3.63\\
        \bottomrule
    \end{tabular}
    }
    \caption{Quantitative evaluation of semantic-to-lidar generation. All models are trained using the SemanticKITTI training partition. The results shown here are adapted from \cite{ran2024towards}. We employ the exact same evaluation toolkit to ensure consistency and comparability.}
    \label{tab:semantic_map_to_lidar}
\end{table*}

\begin{table*}[!th]
    \centering
    \resizebox{\linewidth}{!}{%
        \large
        \begin{tabular}{@{}l l|ccccccccccccccccccc|cc@{}}
            \toprule
             & Methods                & \rotatebox{90}{car} & \rotatebox{90}{bi.cle} & \rotatebox{90}{mt.cle} & \rotatebox{90}{truck} & \rotatebox{90}{oth-v.} & \rotatebox{90}{pers.} & \rotatebox{90}{bi.clst} & \rotatebox{90}{mt.clst} & \rotatebox{90}{road} & \rotatebox{90}{parki.} & \rotatebox{90}{sidew.} & \rotatebox{90}{other-g.} & \rotatebox{90}{build.} & \rotatebox{90}{fence} & \rotatebox{90}{veget.} & \rotatebox{90}{trunk} & \rotatebox{90}{terr.} & \rotatebox{90}{pole} & \rotatebox{90}{traf.} & \rotatebox{90}{mIoU} & \rotatebox{90}{gain} \\
            \midrule
            {} & Source Only          & {95.7}   & {25.0}   & {57.0}   & {62.1}   & {46.4}   & {63.4}   & {77.3}   & {0.0}    & {93.0}   & {47.9}   & {80.5}   & {2.2}    & {89.7}   & {58.6}   & {\underline{89.5}}   & {66.5}   & {\underline{78.0}}   & {64.6}   & {50.1}   & {60.3}   & {+0.0} \\
            \midrule
            \multirow{3}{*}{A} 
              & Jittering \cite{qi2017pointnet} 
                  & {95.7}   & {27.8}   & {56.2}   & {66.0}   & {45.8}   & {65.3}   & {82.8}   & {0.0}    & {93.0}   & {48.2}   & {79.9}   & {\underline{2.5}}   & {89.7}   & {\textbf{62.9}}   & {88.9}   & {64.0}   & {77.0}   & {64.8}   & {\textbf{51.0}}   & {61.2}   & {+0.9} \\
              & Dropout \cite{srivastava2014dropout}
                  & {96.0}   & {28.5}   & {57.1}   & {65.1}   & {46.4}   & {64.1}   & {83.6}   & {0.1}    & {93.5}   & {47.6}   & {80.1}   & {2.3}    & {89.3}   & {61.9}   & {\textbf{90.1}}   & {\underline{66.9}}   & {\textbf{78.8}}   & {\textbf{65.8}}   & {49.1}   & {61.4}   & {+1.1} \\
              & PointAug \cite{li2020pointaugment}
                  & {95.9}   & {29.2}   & {\underline{70.0}}   & {76.3}   & {50.0}   & {67.0}   & {84.4}   & {\textbf{2.4}}   & {93.8}   & {48.1}   & {\underline{81.2}}   & {\textbf{4.6}}   & {89.8}   & {58.4}   & {87.5}   & {65.4}   & {72.7}   & {62.4}   & {\underline{50.5}}   & {62.6}   & {+2.3} \\
            \midrule
            \multirow{3}{*}{B} 
              & +SynLiDAR \cite{xiao2022transfer}
                  & {95.9}   & {\underline{33.0}}   & {62.8}   & {78.9}   & {50.2}   & {\underline{71.4}}   & {83.5}   & {0.7}    & {92.3}   & {\underline{52.8}}   & {79.9}   & {0.1}    & {89.8}   & {59.5}   & {86.3}   & {65.4}   & {72.8}   & {63.6}   & {48.9}   & {62.5}   & {+2.2} \\
              & +LiDM \cite{ran2024towards}
                  & {95.5}   & {19.3}   & {50.4}   & {77.8}   & {46.0}   & {65.7}   & {74.2}   & {0.0}    & {93.7}   & {46.4}   & {80.7}   & {0.8}    & {90.0}   & {59.3}   & {87.3}   & {65.3}   & {74.0}   & {62.3}   & {45.8}   & {59.7}   & {-0.6} \\
              & +SG-LDM (ours)
                  & {\underline{96.5}}   & {24.8}   & {52.1}   & {80.0}   & {56.3}   & {67.1}   & {\underline{86.2}}   & {0.0}    & {\textbf{94.4}}   & {47.9}   & {\textbf{81.6}}   & {0.3}    & {\underline{90.9}}   & {\underline{62.7}}   & {87.3}   & {\underline{66.9}}   & {73.4}   & {63.3}   & {48.8}   & {62.1}   & {+1.8} \\
            \midrule
            \multirow{3}{*}{C} 
              & PCT \cite{xiao2022transfer}
                  & {96.3}   & {\textbf{38.7}}   & {\textbf{73.4}}   & {\underline{82.9}}   & {56.1}   & {71.1}   & {85.3}   & {\underline{1.6}}   & {\underline{94.1}}   & {\textbf{54.3}}   & {\textbf{81.6}}   & {1.3}    & {89.5}   & {59.6}   & {87.8}   & {\underline{66.9}}   & {73.6}   & {\underline{65.4}}   & {\underline{50.5}}   & {\textbf{64.7}}   & {\textbf{+4.4}} \\
              & DGT \cite{yuan2024density}
                  & {96.4}   & {30.8}   & {63.6}   & {81.2}   & {\underline{57.5}}   & {71.0}   & {\textbf{86.5}}   & {0.0}    & {\textbf{94.4}}   & {47.4}   & {81.4}   & {2.2}    & {\underline{90.9}}   & {60.8}   & {87.3}   & {\textbf{67.9}}   & {73.4}   & {62.7}   & {48.0}   & {63.3}   & {+3.0} \\
              & SG-LDM + LT (ours)
                  & {\textbf{97.3}}   & {24.0}   & {59.4}   & {\textbf{91.5}}   & {\textbf{73.6}}   & {\textbf{71.7}}   & {84.5}   & {0.0}    & {94.0}   & {45.4}   & {81.0}   & {0.6}    & {\textbf{91.0}}   & {61.3}   & {87.2}   & {65.0}   & {72.9}   & {62.8}   & {46.3}   & {\underline{63.7}}   & {\underline{+3.4}} \\
            \bottomrule
        \end{tabular}%
    }
    \caption{Data Augmentation Results. The baseline lidar segmentation model, MinkovUnet \cite{choy20194d}, is trained on the SemanticKITTI training partition and augmented using various methods. Group A applies augmentation directly to real data. Group B uses synthetic lidar point clouds generated either from a virtual environment (SynLiDAR) or via semantic-to-lidar generative models guided by SynLiDAR semantic labels. Group C combines SynLiDAR point clouds with a lidar translation (LT) method for domain adaptation.}
    \label{tab:Syn2Sk_XYZ_tab}
\end{table*}

\subsection{Experimental Setup}
Our experiments consist of two stages: data generation and data augmentation. We utilize two datasets, SemanticKITTI \cite{behley2019semantickitti} and SynLiDAR \cite{xiao2022transfer}, in both sets of experiments. In the first stage, we evaluate the data generation performance of the proposed diffusion model and compare it with the state-of-the-art, where all the diffusion models are exclusively trained on the official training partition of SemanticKITTI, and tested on the validation partition of SemanticKITTI. We also evaluate the transferability of the models on the test partition of SynLiDAR. For fast sampling, we employ DDIM for both our SG-LDM and LiDM, reducing the original 1000 DDPM steps to 50 steps. The diffusion model is a standard 2D diffusion with the conventional convolution neural networks replaced by circular convolution \cite{schubert2019circular}.

In the second stage, we use MinkovUnet \cite{choy20194d} as the baseline segmentation model to evaluate the performance of models trained with different data augmentation methods. These methods include traditional techniques such as jittering and random drop, synthetic data generated from virtual environments or generative models, and domain translation applied to the synthetic data produced in virtual environments. All experiments, encompassing both generative and segmentation models, are conducted using four Nvidia A100 GPUs.

\subsection{Experiments on Data Generation}
In this section, we evaluate the generated data against the ground truth. Following \cite{ran2024towards}, we employ FRID, FSVD, and FPVD as perceptual metrics and JSD and MMD as statistical metrics. FRID, FSVD, and FPVD are analogous to the FID score \cite{heusel2017gans} used in image generation. They are computed using different point cloud representation learning backbones, namely, RangeNet++ \cite{milioto2019rangenet++}, MinkowskiNet \cite{choy20194d}, and SPVCNN \cite{tang2020searching}. JSD measures the diversity of the marginal distribution of two sets of point clouds, while MMD calculates the average distance between matched point clouds, indicating fidelity.

\Cref{tab:semantic_map_to_lidar} presents the quantitative results, demonstrating that our method establishes a new state-of-the-art for semantic-to-lidar generation on most metrics. Specifically, when evaluated in the same domain (i.e., testing on SemanticKITTI), our approach outperforms the previous state-of-the-art method, LiDM \cite{ran2024towards}, by $48.0\% \sim 80.8\%$ in the perceptual metrics and achieves a 59\% improvement in MMD. Although our method sacrifices 17\% performance in the JSD score due to its reliance on classifier-free guidance to balance fidelity and diversity, it still outperforms alternative approaches overall. The evaluation in a different domain (SynLiDAR) reveals that our method continues to outperform LiDM by $35.6\% \sim 47.3\%$ in the perceptual metrics. LiDM appears to perform slighlty better in terms of MMD. However, given that SynLiDAR exhibits a significantly different point range, LiDM produces a significant amount of noisy points (as shown in \Cref{fig:teaser}) which is not captured by the statistical metrics.

\begin{figure*}[!thbp]
\centering 
\includegraphics[trim=0.0cm 0.7cm 0.0cm 0.0cm, width=0.9\textwidth]{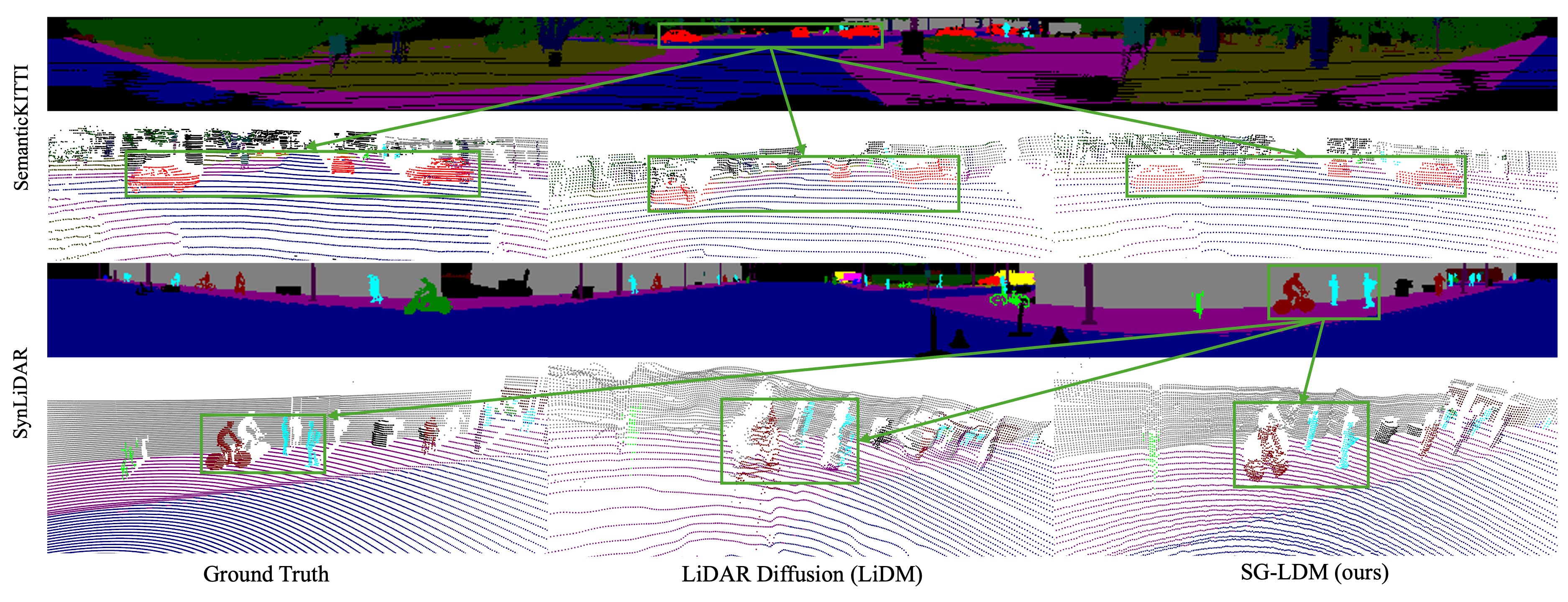}
\caption{Qualitative comparison of generated lidar point clouds between the LiDM and our SG-LDM.}   
\label{fig:Qualitative Evaluation}
\end{figure*}

\Cref{fig:Qualitative Evaluation} further zooms in on object details and demonstrates the correspondence between semantic labels and the generated lidar point clouds. LiDM exhibits significant object collapse in the generated SemanticKITTI point clouds, whereas our method preserves the structural integrity of objects. As for SynLiDAR, LiDM suffers from considerable degradation in object and ground point cloud quality. This is attributed to the VAE used for latent diffusion being unable to generalize to lidar data sourced from a different domain.

\subsection{Experiments on Data Augmentation}
In this section, we leverage the official SemanticKITTI training and validation datasets to evaluate the performance of a baseline segmentation model under various data augmentation strategies, which we categorize into three groups. Group A comprises augmentation techniques applied directly to the original real data. Group B augments the dataset using synthetic lidar point clouds generated either by a virtual environment (SynLiDAR) or by semantic-to-lidar generative models, with the latter leveraging semantic labels from the SynLiDAR dataset as guidance. Group C employs SynLiDAR point clouds in combination with a lidar translation method to facilitate domain adaptation.

\Cref{tab:Syn2Sk_XYZ_tab} presents the results from all groups. In Group B, the data generated by SG-LDM demonstrates a clear data augmentation effect. When using SG-LDM–generated data, the performance is only 0.4\% lower than with SynLiDAR, demonstrating that relying solely on semantic information can yield comparable results. In contrast, due to the poor quality of the data generated by LiDM, incorporating it as additional training data actually degrades model performance.

In Group C, we evaluate the data augmentation performance using synthetic data from SynLiDAR with various lidar transation methods. As a density translation method, our approach shows strong potential by outperforming the state-of-the-art density guided translator (DGT) \cite{yuan2024density}. However, since our method focuses solely on point cloud density, there remains some gap between our approach and PCT \cite{xiao2022transfer}, which considers both the density and appearance of point clouds.



\section{Ablation Study}
\subsection{Semantic Alignment} \label{sec:ablation1}
In this section, we compare three variations of the proposed conditional diffusion model: one without CFG, one with regular CFG, and one with CFG combined with semantic alignment (SA). Both experiments with CFG are conducted with $P_{uncon}=0.2$ and a CFG scale of 2. \Cref{tab:ablation 1} presents the quantitative results for these three models. We observe a clear drop in performance when CFG is applied to a conditional model without the semantic alignment module. However, when semantic alignment is incorporated, classifier-free guidance maintains its intended behavior, which successfully trading off fidelity, as indicated by the perceptual metrics and MMD, with only a minimal cost in diversity as measured by JSD.
\begin{table}[ht]
    \centering
    \resizebox{\linewidth}{!}{
    \large
    \begin{tabular}{l|ccccc}
        \toprule
        \textbf{Method} 
        & \textbf{FRID} $\downarrow$ 
        & \textbf{FSVD} $\downarrow$ 
        & \textbf{FPVD} $\downarrow$ 
        & \textbf{JSD} $\downarrow$ 
        & \textbf{MMD} ($\times10^{-4}$) $\downarrow$ \\
        \midrule
        No CFG & 9.4 & 10.8 & 9.3 & \textbf{0.078} & 1.43\\
        CFG w/o SA & 10.4 & 32.1 &24.7 &0.222 & 1.75\\
        CFG w/ SA & \textbf{4.4} & \textbf{10.5} & \textbf{7.9} & 0.084 & \textbf{1.31}\\
        \bottomrule
    \end{tabular}
    }
    \caption{Evaluation of the different components of SG-LDM on SemanticKITTI. The CFG scale is set as 2, $\mathbf{P}_{uncon}$ is set as 20\%.}
    \label{tab:ablation 1}
\end{table}

\subsection{Classifier-Free Guidance}
In this section, we present an analysis of two key parameters related to classifier-free guidance: \(\mathbf{P}_{\text{uncon}}\) and the CFG scale. \Cref{tab:ablation 2} presents the quantitative analysis of the performance of our SG-LDM under different \(P_{\text{uncon}}\) values. The results are quite stable across various settings, except for a notable drop in the FRID score when \(P_{\text{uncon}}\) is set to 0.5. We attribute this decline to the unconditional generation component beginning to dominate the training process, which leads to a degradation in conditional generation performance. When \(P_{\text{uncon}}\) is too high, the model's focus shifts away from the conditioning information, thereby compromising the quality of the generated data.
\begin{table}[ht]
    \centering
    \resizebox{\linewidth}{!}{
    \large
    \begin{tabular}{l|ccccc}
        \toprule
        \textbf{$P_{uncon}$} 
        & \textbf{FRID} $\downarrow$ 
        & \textbf{FSVD} $\downarrow$ 
        & \textbf{FPVD} $\downarrow$ 
        & \textbf{JSD} $\downarrow$ 
        & \textbf{MMD} ($\times10^{-4}$) $\downarrow$ \\
        \midrule
        0.1 & 4.5 & 11.9 & 9.27 & 0.085 & \textbf{1.26} \\
        0.2 & \textbf{4.4} & \textbf{10.5} & 7.9 & \textbf{0.084} & 1.31\\
        0.5 & 14.5 & 9.5 & \textbf{7.8} & 0.099 & 1.75 \\
        \bottomrule
    \end{tabular}
    }
    \caption{Effect of $\mathbf{P}_{uncon}$ on semantic-to-lidar generation using SemanticKITTI. The CFG scale is set as 2.}
    \label{tab:ablation 2}
\end{table}

\begin{figure}[htp] 
    \centering
        \includegraphics[width=0.4\textwidth]{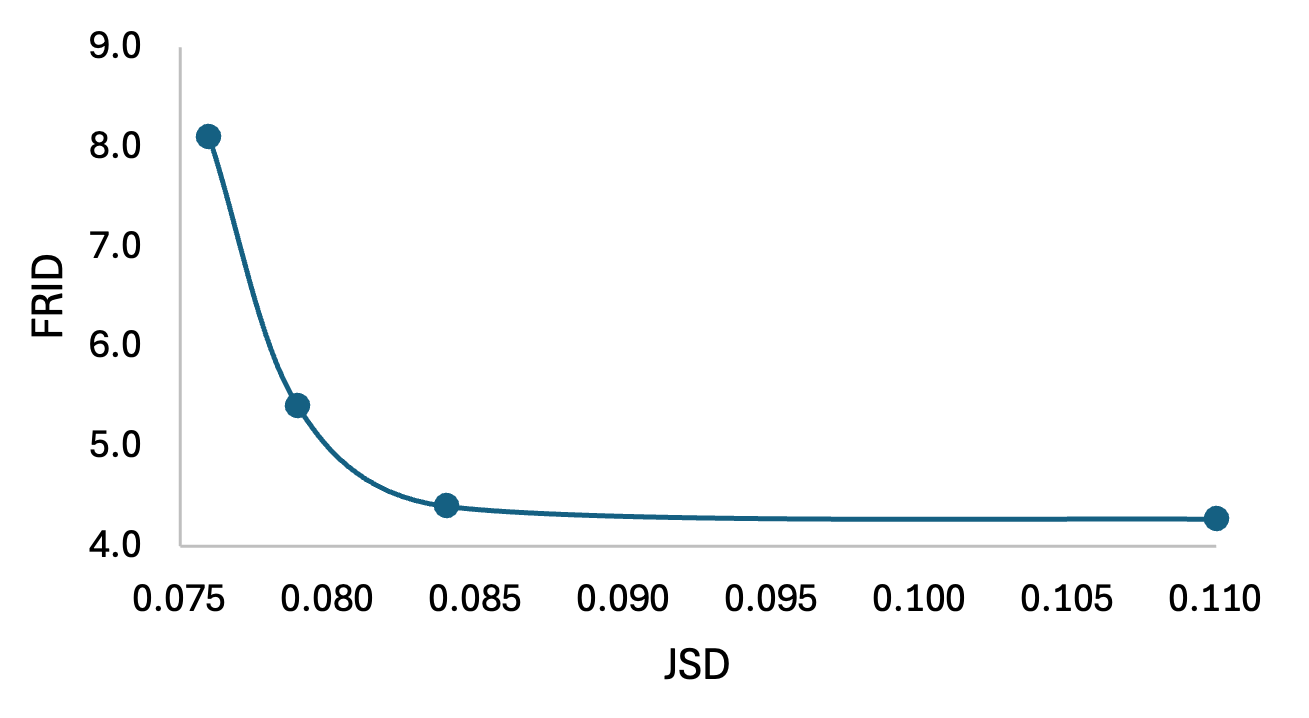}
    \caption{Trade-off between FRID (fidelity) and JSD (diversity). Data points correspond to CFG scales of 1.1, 1.5, 2.0, and 4.0, respectively.}
    \vspace{-1.0em}
    \label{fig: ablation2}
\end{figure}

\Cref{fig: ablation2} illustrates the trade-off between fidelity and diversity for CFG scales ranging from 1.1 to 4.0. We use FRID as the fidelity metric because our point clouds undergo a range conversion process, ensuring that only points which can be projected to a range image and re-projected back are evaluated. As shown in the figure, we successfully balance fidelity and diversity, with FRID achieving optimal performance at a CFG scale of 2.

\subsection{Noising Percentage in Lidar Translation}
\begin{figure}[!thbp]
\vspace{-0.4em}
\centering 
\includegraphics[width=0.8\linewidth]{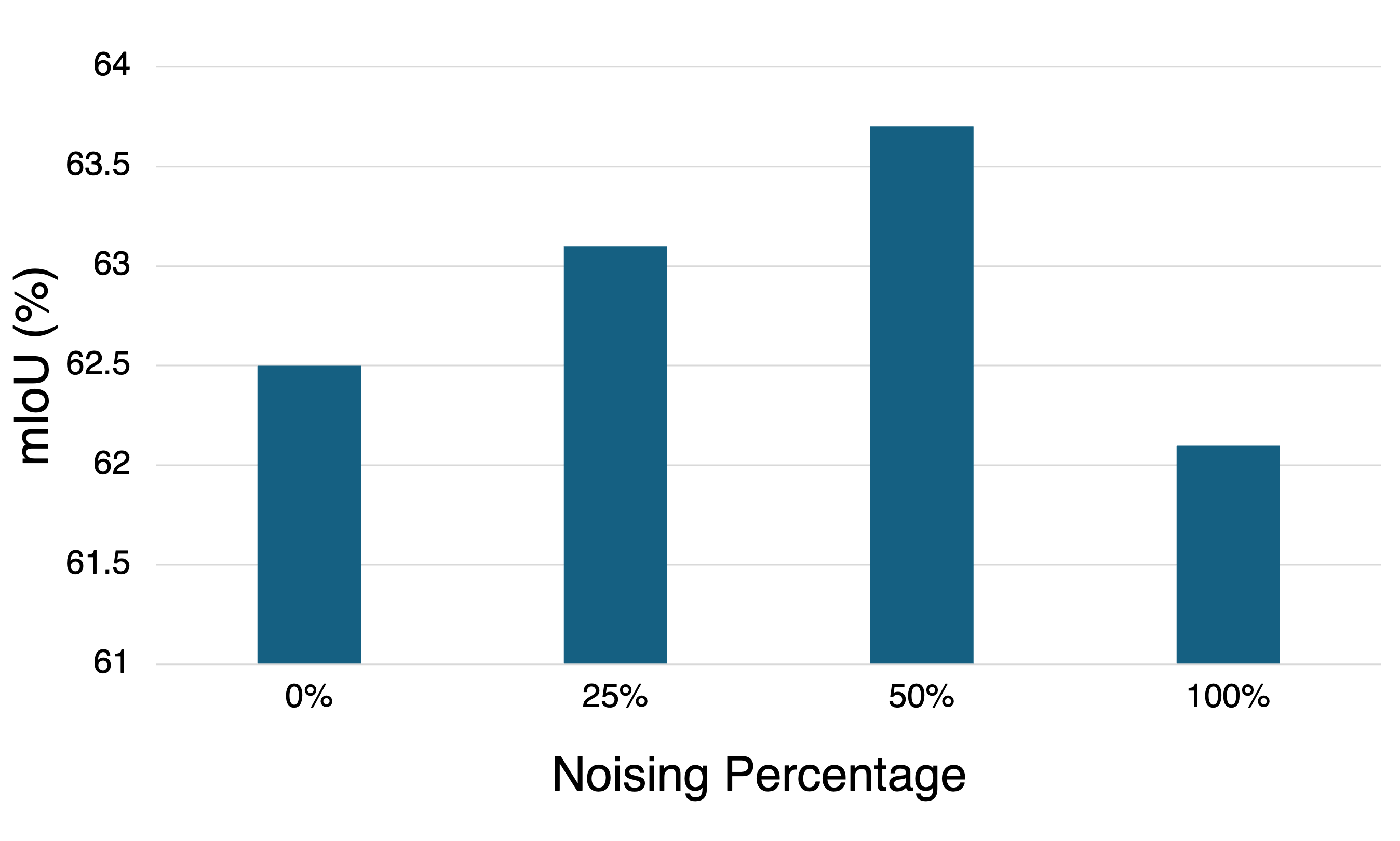}
\vspace{-1.0em}
\caption{Quantitative Comparison  of the data augmentation performance under different noising percentage in the lidar translation framework. If we have 1000 DDPM steps, 50\% denoising means applying forward diffusion for 500 steps on the source data and then performing 500 backward diffusion steps using SG-LDM.}   
\label{fig:ablation 3}
\vspace{-1.0em}
\end{figure}

We conduct an ablation study on the noising percentage applied to the source data in the lidar translation framework. Here, 0\% represents using the raw data from SynLiDAR without any translation, while 100\% indicates that the data is fully generated by SG-LDM. As shown in \Cref{fig:ablation 3}, the best performance is achieved at 50\%. This observation suggests that a moderate amount of noise introduced during the generation process helps the model produce augmented data that is both realistic and diverse. At 50\%, the process strikes an effective balance by preserving the basic geometric structure from the synthetic dataset while also incorporating the translation benefits learned by SG-LDM from real data.

\subsection{Efficiency}
One of the main challenges of traditional diffusion models is their inference-time efficiency. In our work, we benchmark SG-LDM against a latent diffusion model, LiDM \cite{ran2024towards}, using a single Nvidia A100 GPU. As shown in \Cref{tab:efficiency}, while LiDM is 170\% faster at inference, its performance on the FRID metric is 420.5\% inferior compared to SG-LDM. These results suggest that the additional inference cost of SG-LDM is a worthwhile trade-off for its significant performance improvements.

\begin{table}[ht]
    \centering
    \resizebox{\linewidth}{!}{
    \large
    \begin{tabular}{l|ccc}
        \toprule
        \textbf{Method} 
        & \textbf{Latent Diffusion} 
        & \textbf{Throughput} $\uparrow$ 
        & \textbf{Infer. Speed} $\uparrow$ \\ 
        \midrule
        LiDM \cite{ran2024towards} &\cmark & \textbf{1.711}  & \textbf{85.53} \\
        SG-LDM (ours) &\xmark & 0.634 & 31.68\\
        \bottomrule
    \end{tabular}
    }
    \caption{Comparison on efficiency of LiDM and our SG-LDM using the semanticKITTI validation set. We test both models on one NVIDIA A100 with different batch sizes to ensure full utilization of 80GB GPU memory. Both applied DDIM at the inference stage. Throughput is defined as the number of generated samples per second, while inference speed refers to the number of diffusion steps executed per second.}
    \vspace{-1.0em}
    \label{tab:efficiency}
\end{table}
\section{Limitations and Future Work}
The primary limitation of our lidar translation framework is that we can only transfer density, not appearance. This constraint arises because the translation process is guided by a semantic label that is generated from a virtual environment. As a result, our method does not outperform the state-of-the-art method PCT \cite{xiao2022transfer}, which translate both the density and appearance using conditional GAN. Future work should explore ways to relax this requirement, either by using an alternative guidance mechanism or by first translating the semantic map to better match the appearance of lidar point clouds. Additionally, our approach relies on the standard DDPM architecture, which is less efficient than LiDM \cite{ran2024towards}. Future research should focus on developing methods for training VAEs that are robust to the domain variability in lidar point clouds, thereby laying the foundation for an effective latent diffusion model for lidar data.

\section{Conclusion}
In this paper we proposed SG-LDM, a novel lidar diffusion model specifically designed for the semantic-to-lidar task. We evaluated our approach under two settings: by comparing the quality of the generated data and by using the generated data to augment a semantic segmentation model. In both cases, our method clearly outperforms the previous state of the art. Furthermore, we introduce a simple yet effective semantic-guided lidar translation framework that achieves comparable performance and a more stable alternative to GAN-based approaches, surpassing synthetic data directly generated from virtual environments.
\section*{Acknowledgments}
\noindent The first two authors acknowledge the financial support from The University of Melbourne through the Melbourne Research Scholarship. This research was supported by The University of Melbourne’s Research Computing Services and the Petascale Campus Initiative.
{
    \small
    \bibliographystyle{ieeenat_fullname}
    \bibliography{main}

\begin{thebibliography}{77}
\providecommand{\natexlab}[1]{#1}
\providecommand{\url}[1]{\texttt{#1}}
\expandafter\ifx\csname urlstyle\endcsname\relax
  \providecommand{\doi}[1]{doi: #1}\else
  \providecommand{\doi}{doi: \begingroup \urlstyle{rm}\Url}\fi

\bibitem[Acharya et~al.(2023)Acharya, Tatli, and
  Khoshelham]{acharya2023synthetic}
Debaditya Acharya, Christopher~James Tatli, and Kourosh Khoshelham.
\newblock Synthetic-real image domain adaptation for indoor camera pose
  regression using a 3d model.
\newblock \emph{ISPRS Journal of Photogrammetry and Remote Sensing},
  202:\penalty0 405--421, 2023.

\bibitem[Achlioptas et~al.(2018)Achlioptas, Diamanti, Mitliagkas, and
  Guibas]{achlioptas2018learning}
Panos Achlioptas, Olga Diamanti, Ioannis Mitliagkas, and Leonidas Guibas.
\newblock Learning representations and generative models for 3d point clouds.
\newblock In \emph{International conference on machine learning}, pages 40--49.
  PMLR, 2018.

\bibitem[Behley et~al.(2019)Behley, Garbade, Milioto, Quenzel, Behnke,
  Stachniss, and Gall]{behley2019semantickitti}
Jens Behley, Martin Garbade, Andres Milioto, Jan Quenzel, Sven Behnke, Cyrill
  Stachniss, and Jurgen Gall.
\newblock Semantickitti: A dataset for semantic scene understanding of lidar
  sequences.
\newblock In \emph{Proceedings of the IEEE/CVF international conference on
  computer vision}, pages 9297--9307, 2019.

\bibitem[Blattmann et~al.(2023)Blattmann, Rombach, Ling, Dockhorn, Kim, Fidler,
  and Kreis]{blattmann2023align}
Andreas Blattmann, Robin Rombach, Huan Ling, Tim Dockhorn, Seung~Wook Kim,
  Sanja Fidler, and Karsten Kreis.
\newblock Align your latents: High-resolution video synthesis with latent
  diffusion models.
\newblock In \emph{Proceedings of the IEEE/CVF conference on computer vision
  and pattern recognition}, pages 22563--22575, 2023.

\bibitem[Caccia et~al.(2019)Caccia, Van~Hoof, Courville, and
  Pineau]{caccia2019deep}
Lucas Caccia, Herke Van~Hoof, Aaron Courville, and Joelle Pineau.
\newblock Deep generative modeling of lidar data.
\newblock In \emph{2019 IEEE/RSJ International Conference on Intelligent Robots
  and Systems (IROS)}, pages 5034--5040. IEEE, 2019.

\bibitem[Caesar et~al.(2020)Caesar, Bankiti, Lang, Vora, Liong, Xu, Krishnan,
  Pan, Baldan, and Beijbom]{caesar2020nuscenes}
Holger Caesar, Varun Bankiti, Alex~H Lang, Sourabh Vora, Venice~Erin Liong,
  Qiang Xu, Anush Krishnan, Yu Pan, Giancarlo Baldan, and Oscar Beijbom.
\newblock nuscenes: A multimodal dataset for autonomous driving.
\newblock In \emph{Proceedings of the IEEE/CVF conference on computer vision
  and pattern recognition}, pages 11621--11631, 2020.

\bibitem[Choi et~al.(2018)Choi, Choi, Kim, Ha, Kim, and Choo]{choi2018stargan}
Yunjey Choi, Minje Choi, Munyoung Kim, Jung-Woo Ha, Sunghun Kim, and Jaegul
  Choo.
\newblock Stargan: Unified generative adversarial networks for multi-domain
  image-to-image translation.
\newblock In \emph{Proceedings of the IEEE conference on computer vision and
  pattern recognition}, pages 8789--8797, 2018.

\bibitem[Choy et~al.(2019)Choy, Gwak, and Savarese]{choy20194d}
Christopher Choy, JunYoung Gwak, and Silvio Savarese.
\newblock 4d spatio-temporal convnets: Minkowski convolutional neural networks.
\newblock In \emph{Proceedings of the IEEE/CVF conference on computer vision
  and pattern recognition}, pages 3075--3084, 2019.

\bibitem[Dhariwal and Nichol(2021)]{dhariwal2021diffusion}
Prafulla Dhariwal and Alexander Nichol.
\newblock Diffusion models beat gans on image synthesis.
\newblock \emph{Advances in neural information processing systems},
  34:\penalty0 8780--8794, 2021.

\bibitem[Dosovitskiy et~al.(2017)Dosovitskiy, Ros, Codevilla, Lopez, and
  Koltun]{dosovitskiy2017carla}
Alexey Dosovitskiy, German Ros, Felipe Codevilla, Antonio Lopez, and Vladlen
  Koltun.
\newblock Carla: An open urban driving simulator.
\newblock In \emph{Conference on robot learning}, pages 1--16. PMLR, 2017.

\bibitem[Esser et~al.(2021)Esser, Rombach, and Ommer]{esser2021taming}
Patrick Esser, Robin Rombach, and Bjorn Ommer.
\newblock Taming transformers for high-resolution image synthesis.
\newblock In \emph{Proceedings of the IEEE/CVF conference on computer vision
  and pattern recognition}, pages 12873--12883, 2021.

\bibitem[Esser et~al.(2023)Esser, Chiu, Atighehchian, Granskog, and
  Germanidis]{esser2023structure}
Patrick Esser, Johnathan Chiu, Parmida Atighehchian, Jonathan Granskog, and
  Anastasis Germanidis.
\newblock Structure and content-guided video synthesis with diffusion models.
\newblock In \emph{Proceedings of the IEEE/CVF international conference on
  computer vision}, pages 7346--7356, 2023.

\bibitem[Gal et~al.(2022)Gal, Patashnik, Maron, Bermano, Chechik, and
  Cohen-Or]{gal2022stylegan}
Rinon Gal, Or Patashnik, Haggai Maron, Amit~H Bermano, Gal Chechik, and Daniel
  Cohen-Or.
\newblock Stylegan-nada: Clip-guided domain adaptation of image generators.
\newblock \emph{ACM Transactions on Graphics (TOG)}, 41\penalty0 (4):\penalty0
  1--13, 2022.

\bibitem[Heusel et~al.(2017)Heusel, Ramsauer, Unterthiner, Nessler, and
  Hochreiter]{heusel2017gans}
Martin Heusel, Hubert Ramsauer, Thomas Unterthiner, Bernhard Nessler, and Sepp
  Hochreiter.
\newblock Gans trained by a two time-scale update rule converge to a local nash
  equilibrium.
\newblock \emph{Advances in neural information processing systems}, 30, 2017.

\bibitem[Ho and Salimans(2022)]{ho2022classifier}
Jonathan Ho and Tim Salimans.
\newblock Classifier-free diffusion guidance.
\newblock \emph{arXiv preprint arXiv:2207.12598}, 2022.

\bibitem[Ho et~al.(2020)Ho, Jain, and Abbeel]{ho2020denoising}
Jonathan Ho, Ajay Jain, and Pieter Abbeel.
\newblock Denoising diffusion probabilistic models.
\newblock \emph{Advances in neural information processing systems},
  33:\penalty0 6840--6851, 2020.

\bibitem[Ho et~al.(2022{\natexlab{a}})Ho, Chan, Saharia, Whang, Gao, Gritsenko,
  Kingma, Poole, Norouzi, Fleet, and Salimans]{ho2022imagenvideohighdefinition}
Jonathan Ho, William Chan, Chitwan Saharia, Jay Whang, Ruiqi Gao, Alexey
  Gritsenko, Diederik~P. Kingma, Ben Poole, Mohammad Norouzi, David~J. Fleet,
  and Tim Salimans.
\newblock Imagen video: High definition video generation with diffusion models,
  2022{\natexlab{a}}.

\bibitem[Ho et~al.(2022{\natexlab{b}})Ho, Saharia, Chan, Fleet, Norouzi, and
  Salimans]{ho2022cascaded}
Jonathan Ho, Chitwan Saharia, William Chan, David~J Fleet, Mohammad Norouzi,
  and Tim Salimans.
\newblock Cascaded diffusion models for high fidelity image generation.
\newblock \emph{Journal of Machine Learning Research}, 23\penalty0
  (47):\penalty0 1--33, 2022{\natexlab{b}}.

\bibitem[Ho et~al.(2022{\natexlab{c}})Ho, Salimans, Gritsenko, Chan, Norouzi,
  and Fleet]{ho2022video}
Jonathan Ho, Tim Salimans, Alexey Gritsenko, William Chan, Mohammad Norouzi,
  and David~J Fleet.
\newblock Video diffusion models.
\newblock \emph{Advances in Neural Information Processing Systems},
  35:\penalty0 8633--8646, 2022{\natexlab{c}}.

\bibitem[Hu et~al.(2024)Hu, Zhang, and Hu]{hu2024rangeldm}
Qianjiang Hu, Zhimin Zhang, and Wei Hu.
\newblock Rangeldm: Fast realistic lidar point cloud generation.
\newblock In \emph{European Conference on Computer Vision}, pages 115--135.
  Springer, 2024.

\bibitem[Jun and Nichol(2023)]{jun2023shap}
Heewoo Jun and Alex Nichol.
\newblock Shap-e: Generating conditional 3d implicit functions.
\newblock \emph{arXiv preprint arXiv:2305.02463}, 2023.

\bibitem[Karras et~al.(2019)Karras, Laine, and Aila]{karras2019style}
Tero Karras, Samuli Laine, and Timo Aila.
\newblock A style-based generator architecture for generative adversarial
  networks.
\newblock In \emph{Proceedings of the IEEE/CVF conference on computer vision
  and pattern recognition}, pages 4401--4410, 2019.

\bibitem[Kong et~al.(2023)Kong, Liu, Chen, Ma, Zhu, Li, Hou, Qiao, and
  Liu]{kong2023rethinking}
Lingdong Kong, Youquan Liu, Runnan Chen, Yuexin Ma, Xinge Zhu, Yikang Li,
  Yuenan Hou, Yu Qiao, and Ziwei Liu.
\newblock Rethinking range view representation for lidar segmentation.
\newblock In \emph{Proceedings of the IEEE/CVF International Conference on
  Computer Vision}, pages 228--240, 2023.

\bibitem[Lai et~al.(2023)Lai, Chen, Lu, Liu, and Jia]{lai2023spherical}
Xin Lai, Yukang Chen, Fanbin Lu, Jianhui Liu, and Jiaya Jia.
\newblock Spherical transformer for lidar-based 3d recognition.
\newblock In \emph{Proceedings of the IEEE/CVF Conference on Computer Vision
  and Pattern Recognition}, pages 17545--17555, 2023.

\bibitem[Lao et~al.(2025)Lao, Tang, Wu, Chen, Yu, and Zhao]{lao2025lit}
Yixing Lao, Tao Tang, Xiaoyang Wu, Peng Chen, Kaicheng Yu, and Hengshuang Zhao.
\newblock Lit: Unifying lidar" languages" with lidar translator.
\newblock \emph{Advances in Neural Information Processing Systems},
  37:\penalty0 93767--93789, 2025.

\bibitem[Levering et~al.(2020)Levering, Tomko, Tuia, and
  Khoshelham]{levering2020detecting}
Alex Levering, Martin Tomko, Devis Tuia, and Kourosh Khoshelham.
\newblock Detecting unsigned physical road incidents from driver-view images.
\newblock \emph{IEEE Transactions on Intelligent Vehicles}, 6\penalty0
  (1):\penalty0 24--33, 2020.

\bibitem[Li et~al.(2020)Li, Li, Heng, and Fu]{li2020pointaugment}
Ruihui Li, Xianzhi Li, Pheng-Ann Heng, and Chi-Wing Fu.
\newblock Pointaugment: an auto-augmentation framework for point cloud
  classification.
\newblock In \emph{Proceedings of the IEEE/CVF conference on computer vision
  and pattern recognition}, pages 6378--6387, 2020.

\bibitem[Lin et~al.(2023)Lin, Gao, Tang, Takikawa, Zeng, Huang, Kreis, Fidler,
  Liu, and Lin]{lin2023magic3d}
Chen-Hsuan Lin, Jun Gao, Luming Tang, Towaki Takikawa, Xiaohui Zeng, Xun Huang,
  Karsten Kreis, Sanja Fidler, Ming-Yu Liu, and Tsung-Yi Lin.
\newblock Magic3d: High-resolution text-to-3d content creation.
\newblock In \emph{Proceedings of the IEEE/CVF conference on computer vision
  and pattern recognition}, pages 300--309, 2023.

\bibitem[Luo et~al.(2020)Luo, Khoshelham, Fang, and Chen]{luo2020unsupervised}
Haifeng Luo, Kourosh Khoshelham, Lina Fang, and Chongcheng Chen.
\newblock Unsupervised scene adaptation for semantic segmentation of urban
  mobile laser scanning point clouds.
\newblock \emph{ISPRS Journal of Photogrammetry and Remote Sensing},
  169:\penalty0 253--267, 2020.

\bibitem[Luo and Hu(2021)]{luo2021diffusion}
Shitong Luo and Wei Hu.
\newblock Diffusion probabilistic models for 3d point cloud generation.
\newblock In \emph{Proceedings of the IEEE/CVF conference on computer vision
  and pattern recognition}, pages 2837--2845, 2021.

\bibitem[Luo et~al.(2019)Luo, Zheng, Guan, Yu, and Yang]{luo2019taking}
Yawei Luo, Liang Zheng, Tao Guan, Junqing Yu, and Yi Yang.
\newblock Taking a closer look at domain shift: Category-level adversaries for
  semantics consistent domain adaptation.
\newblock In \emph{Proceedings of the IEEE/CVF conference on computer vision
  and pattern recognition}, pages 2507--2516, 2019.

\bibitem[Melas-Kyriazi et~al.(2023)Melas-Kyriazi, Rupprecht, and
  Vedaldi]{melas2023pc2}
Luke Melas-Kyriazi, Christian Rupprecht, and Andrea Vedaldi.
\newblock Pc2: Projection-conditioned point cloud diffusion for single-image 3d
  reconstruction.
\newblock In \emph{Proceedings of the IEEE/CVF Conference on Computer Vision
  and Pattern Recognition}, pages 12923--12932, 2023.

\bibitem[Milioto et~al.(2019)Milioto, Vizzo, Behley, and
  Stachniss]{milioto2019rangenet++}
Andres Milioto, Ignacio Vizzo, Jens Behley, and Cyrill Stachniss.
\newblock Rangenet++: Fast and accurate lidar semantic segmentation.
\newblock In \emph{2019 IEEE/RSJ international conference on intelligent robots
  and systems (IROS)}, pages 4213--4220. IEEE, 2019.

\bibitem[Mo et~al.(2023)Mo, Xie, Chu, Hong, Niessner, and Li]{mo2023dit}
Shentong Mo, Enze Xie, Ruihang Chu, Lanqing Hong, Matthias Niessner, and
  Zhenguo Li.
\newblock Dit-3d: Exploring plain diffusion transformers for 3d shape
  generation.
\newblock \emph{Advances in neural information processing systems},
  36:\penalty0 67960--67971, 2023.

\bibitem[Nakashima and Kurazume(2021)]{nakashima2021learning}
Kazuto Nakashima and Ryo Kurazume.
\newblock Learning to drop points for lidar scan synthesis.
\newblock In \emph{2021 IEEE/RSJ International Conference on Intelligent Robots
  and Systems (IROS)}, pages 222--229. IEEE, 2021.

\bibitem[Nakashima and Kurazume(2024)]{nakashima2024lidar}
Kazuto Nakashima and Ryo Kurazume.
\newblock Lidar data synthesis with denoising diffusion probabilistic models.
\newblock In \emph{2024 IEEE International Conference on Robotics and
  Automation (ICRA)}, pages 14724--14731. IEEE, 2024.

\bibitem[Nakashima et~al.(2023)Nakashima, Iwashita, and
  Kurazume]{nakashima2023generative}
Kazuto Nakashima, Yumi Iwashita, and Ryo Kurazume.
\newblock Generative range imaging for learning scene priors of 3d lidar data.
\newblock In \emph{Proceedings of the IEEE/CVF Winter Conference on
  Applications of Computer Vision}, pages 1256--1266, 2023.

\bibitem[Nichol et~al.(2022{\natexlab{a}})Nichol, Jun, Dhariwal, Mishkin, and
  Chen]{nichol2022point}
Alex Nichol, Heewoo Jun, Prafulla Dhariwal, Pamela Mishkin, and Mark Chen.
\newblock Point-e: A system for generating 3d point clouds from complex
  prompts.
\newblock \emph{arXiv preprint arXiv:2212.08751}, 2022{\natexlab{a}}.

\bibitem[Nichol et~al.(2022{\natexlab{b}})Nichol, Dhariwal, Ramesh, Shyam,
  Mishkin, Mcgrew, Sutskever, and Chen]{nichol2022glide}
Alexander~Quinn Nichol, Prafulla Dhariwal, Aditya Ramesh, Pranav Shyam, Pamela
  Mishkin, Bob Mcgrew, Ilya Sutskever, and Mark Chen.
\newblock Glide: Towards photorealistic image generation and editing with
  text-guided diffusion models.
\newblock In \emph{International Conference on Machine Learning}, pages
  16784--16804. PMLR, 2022{\natexlab{b}}.

\bibitem[Peebles and Xie(2023)]{peebles2023scalable}
William Peebles and Saining Xie.
\newblock Scalable diffusion models with transformers.
\newblock In \emph{Proceedings of the IEEE/CVF international conference on
  computer vision}, pages 4195--4205, 2023.

\bibitem[Peng et~al.(2023)Peng, Hu, Ke, and Liu]{peng2023diffusion}
Duo Peng, Ping Hu, Qiuhong Ke, and Jun Liu.
\newblock Diffusion-based image translation with label guidance for domain
  adaptive semantic segmentation.
\newblock In \emph{Proceedings of the IEEE/CVF international conference on
  computer vision}, pages 808--820, 2023.

\bibitem[Poole et~al.(2023)Poole, Jain, Barron, and
  Mildenhall]{poole2023dreamfusion}
Ben Poole, Ajay Jain, Jonathan~T. Barron, and Ben Mildenhall.
\newblock Dreamfusion: Text-to-3d using 2d diffusion.
\newblock In \emph{The Eleventh International Conference on Learning
  Representations}, 2023.

\bibitem[Qi et~al.(2017)Qi, Su, Mo, and Guibas]{qi2017pointnet}
Charles~R Qi, Hao Su, Kaichun Mo, and Leonidas~J Guibas.
\newblock Pointnet: Deep learning on point sets for 3d classification and
  segmentation.
\newblock In \emph{Proceedings of the IEEE conference on computer vision and
  pattern recognition}, pages 652--660, 2017.

\bibitem[Ramesh et~al.(2022)Ramesh, Dhariwal, Nichol, Chu, and
  Chen]{ramesh2022hierarchical}
Aditya Ramesh, Prafulla Dhariwal, Alex Nichol, Casey Chu, and Mark Chen.
\newblock Hierarchical text-conditional image generation with clip latents.
\newblock \emph{arXiv preprint arXiv:2204.06125}, 2022.

\bibitem[Ran et~al.(2024)Ran, Guizilini, and Wang]{ran2024towards}
Haoxi Ran, Vitor Guizilini, and Yue Wang.
\newblock Towards realistic scene generation with lidar diffusion models.
\newblock In \emph{Proceedings of the IEEE/CVF Conference on Computer Vision
  and Pattern Recognition}, pages 14738--14748, 2024.

\bibitem[Rombach et~al.(2022)Rombach, Blattmann, Lorenz, Esser, and
  Ommer]{rombach2022high}
Robin Rombach, Andreas Blattmann, Dominik Lorenz, Patrick Esser, and Bj{\"o}rn
  Ommer.
\newblock High-resolution image synthesis with latent diffusion models.
\newblock In \emph{Proceedings of the IEEE/CVF conference on computer vision
  and pattern recognition}, pages 10684--10695, 2022.

\bibitem[Ronneberger et~al.(2015)Ronneberger, Fischer, and
  Brox]{ronneberger2015u}
Olaf Ronneberger, Philipp Fischer, and Thomas Brox.
\newblock U-net: Convolutional networks for biomedical image segmentation.
\newblock In \emph{Medical image computing and computer-assisted
  intervention--MICCAI 2015: 18th international conference, Munich, Germany,
  October 5-9, 2015, proceedings, part III 18}, pages 234--241. Springer, 2015.

\bibitem[Saharia et~al.(2022)Saharia, Chan, Saxena, Li, Whang, Denton,
  Ghasemipour, Gontijo~Lopes, Karagol~Ayan, Salimans,
  et~al.]{saharia2022photorealistic}
Chitwan Saharia, William Chan, Saurabh Saxena, Lala Li, Jay Whang, Emily~L
  Denton, Kamyar Ghasemipour, Raphael Gontijo~Lopes, Burcu Karagol~Ayan, Tim
  Salimans, et~al.
\newblock Photorealistic text-to-image diffusion models with deep language
  understanding.
\newblock \emph{Advances in neural information processing systems},
  35:\penalty0 36479--36494, 2022.

\bibitem[Saltori et~al.(2022)Saltori, Galasso, Fiameni, Sebe, Ricci, and
  Poiesi]{saltori2022cosmix}
Cristiano Saltori, Fabio Galasso, Giuseppe Fiameni, Nicu Sebe, Elisa Ricci, and
  Fabio Poiesi.
\newblock Cosmix: Compositional semantic mix for domain adaptation in 3d lidar
  segmentation.
\newblock In \emph{European Conference on Computer Vision}, pages 586--602.
  Springer, 2022.

\bibitem[Schubert et~al.(2019)Schubert, Neubert, P{\"o}schmann, and
  Protzel]{schubert2019circular}
Stefan Schubert, Peer Neubert, Johannes P{\"o}schmann, and Peter Protzel.
\newblock Circular convolutional neural networks for panoramic images and laser
  data.
\newblock In \emph{2019 IEEE intelligent vehicles symposium (IV)}, pages
  653--660. IEEE, 2019.

\bibitem[Shah et~al.(2018)Shah, Dey, Lovett, and Kapoor]{shah2018airsim}
Shital Shah, Debadeepta Dey, Chris Lovett, and Ashish Kapoor.
\newblock Airsim: High-fidelity visual and physical simulation for autonomous
  vehicles.
\newblock In \emph{Field and Service Robotics: Results of the 11th
  International Conference}, pages 621--635. Springer, 2018.

\bibitem[Singer et~al.(2023)Singer, Polyak, Hayes, Yin, An, Zhang, Hu, Yang,
  Ashual, Gafni, Parikh, Gupta, and Taigman]{singer2023makeavideo}
Uriel Singer, Adam Polyak, Thomas Hayes, Xi Yin, Jie An, Songyang Zhang, Qiyuan
  Hu, Harry Yang, Oron Ashual, Oran Gafni, Devi Parikh, Sonal Gupta, and Yaniv
  Taigman.
\newblock Make-a-video: Text-to-video generation without text-video data.
\newblock In \emph{The Eleventh International Conference on Learning
  Representations}, 2023.

\bibitem[Song et~al.(2021)Song, Meng, and Ermon]{song2021denoising}
Jiaming Song, Chenlin Meng, and Stefano Ermon.
\newblock Denoising diffusion implicit models.
\newblock In \emph{International Conference on Learning Representations}, 2021.

\bibitem[Srivastava et~al.(2014)Srivastava, Hinton, Krizhevsky, Sutskever, and
  Salakhutdinov]{srivastava2014dropout}
Nitish Srivastava, Geoffrey Hinton, Alex Krizhevsky, Ilya Sutskever, and Ruslan
  Salakhutdinov.
\newblock Dropout: a simple way to prevent neural networks from overfitting.
\newblock \emph{The journal of machine learning research}, 15\penalty0
  (1):\penalty0 1929--1958, 2014.

\bibitem[Tang et~al.(2020)Tang, Liu, Zhao, Lin, Lin, Wang, and
  Han]{tang2020searching}
Haotian Tang, Zhijian Liu, Shengyu Zhao, Yujun Lin, Ji Lin, Hanrui Wang, and
  Song Han.
\newblock Searching efficient 3d architectures with sparse point-voxel
  convolution.
\newblock In \emph{European conference on computer vision}, pages 685--702.
  Springer, 2020.

\bibitem[Tsai et~al.(2018)Tsai, Hung, Schulter, Sohn, Yang, and
  Chandraker]{tsai2018learning}
Yi-Hsuan Tsai, Wei-Chih Hung, Samuel Schulter, Kihyuk Sohn, Ming-Hsuan Yang,
  and Manmohan Chandraker.
\newblock Learning to adapt structured output space for semantic segmentation.
\newblock In \emph{Proceedings of the IEEE conference on computer vision and
  pattern recognition}, pages 7472--7481, 2018.

\bibitem[Vahdat et~al.(2022)Vahdat, Williams, Gojcic, Litany, Fidler, Kreis,
  et~al.]{vahdat2022lion}
Arash Vahdat, Francis Williams, Zan Gojcic, Or Litany, Sanja Fidler, Karsten
  Kreis, et~al.
\newblock Lion: Latent point diffusion models for 3d shape generation.
\newblock \emph{Advances in Neural Information Processing Systems},
  35:\penalty0 10021--10039, 2022.

\bibitem[Vu et~al.(2019)Vu, Jain, Bucher, Cord, and P{\'e}rez]{vu2019advent}
Tuan-Hung Vu, Himalaya Jain, Maxime Bucher, Matthieu Cord, and Patrick
  P{\'e}rez.
\newblock Advent: Adversarial entropy minimization for domain adaptation in
  semantic segmentation.
\newblock In \emph{Proceedings of the IEEE/CVF conference on computer vision
  and pattern recognition}, pages 2517--2526, 2019.

\bibitem[Wang et~al.(2020)Wang, Shen, Zhang, Duan, and Mei]{wang2020classes}
Haoran Wang, Tong Shen, Wei Zhang, Ling-Yu Duan, and Tao Mei.
\newblock Classes matter: A fine-grained adversarial approach to cross-domain
  semantic segmentation.
\newblock In \emph{European conference on computer vision}, pages 642--659.
  Springer, 2020.

\bibitem[Wu et~al.(2023{\natexlab{a}})Wu, Ge, Wang, Lei, Gu, Shi, Hsu, Shan,
  Qie, and Shou]{Wu_2023_ICCV}
Jay~Zhangjie Wu, Yixiao Ge, Xintao Wang, Stan~Weixian Lei, Yuchao Gu, Yufei
  Shi, Wynne Hsu, Ying Shan, Xiaohu Qie, and Mike~Zheng Shou.
\newblock Tune-a-video: One-shot tuning of image diffusion models for
  text-to-video generation.
\newblock In \emph{Proceedings of the IEEE/CVF International Conference on
  Computer Vision (ICCV)}, pages 7623--7633, 2023{\natexlab{a}}.

\bibitem[Wu et~al.(2023{\natexlab{b}})Wu, Wang, Gong, Liu, Xiong, Ranjan,
  Krishnamoorthi, Chandra, and Liu]{wu2023fast}
Lemeng Wu, Dilin Wang, Chengyue Gong, Xingchao Liu, Yunyang Xiong, Rakesh
  Ranjan, Raghuraman Krishnamoorthi, Vikas Chandra, and Qiang Liu.
\newblock Fast point cloud generation with straight flows.
\newblock In \emph{Proceedings of the IEEE/CVF conference on computer vision
  and pattern recognition}, pages 9445--9454, 2023{\natexlab{b}}.

\bibitem[Wu et~al.(2022)Wu, Lao, Jiang, Liu, and Zhao]{wu2022point}
Xiaoyang Wu, Yixing Lao, Li Jiang, Xihui Liu, and Hengshuang Zhao.
\newblock Point transformer v2: Grouped vector attention and partition-based
  pooling.
\newblock \emph{Advances in Neural Information Processing Systems},
  35:\penalty0 33330--33342, 2022.

\bibitem[Wu et~al.(2024{\natexlab{a}})Wu, Jiang, Wang, Liu, Liu, Qiao, Ouyang,
  He, and Zhao]{wu2024point}
Xiaoyang Wu, Li Jiang, Peng-Shuai Wang, Zhijian Liu, Xihui Liu, Yu Qiao, Wanli
  Ouyang, Tong He, and Hengshuang Zhao.
\newblock Point transformer v3: Simpler faster stronger.
\newblock In \emph{Proceedings of the IEEE/CVF Conference on Computer Vision
  and Pattern Recognition}, pages 4840--4851, 2024{\natexlab{a}}.

\bibitem[Wu et~al.(2024{\natexlab{b}})Wu, Tian, Wen, Peng, Liu, Yu, and
  Zhao]{wu2024towards}
Xiaoyang Wu, Zhuotao Tian, Xin Wen, Bohao Peng, Xihui Liu, Kaicheng Yu, and
  Hengshuang Zhao.
\newblock Towards large-scale 3d representation learning with multi-dataset
  point prompt training.
\newblock In \emph{Proceedings of the IEEE/CVF Conference on Computer Vision
  and Pattern Recognition}, pages 19551--19562, 2024{\natexlab{b}}.

\bibitem[Wu et~al.(2024{\natexlab{c}})Wu, Zhang, Qian, Xie, and
  Yang]{wu2024text2lidar}
Yang Wu, Kaihua Zhang, Jianjun Qian, Jin Xie, and Jian Yang.
\newblock Text2lidar: Text-guided lidar point cloud generation via
  equirectangular transformer.
\newblock In \emph{European Conference on Computer Vision}, pages 291--310.
  Springer, 2024{\natexlab{c}}.

\bibitem[Xiang et~al.(2024)Xiang, Huang, and Khoshelham]{xiang2024synthetic}
Zhengkang Xiang, Zexian Huang, and Kourosh Khoshelham.
\newblock Synthetic lidar point cloud generation using deep generative models
  for improved driving scene object recognition.
\newblock \emph{Image and Vision Computing}, 150:\penalty0 105207, 2024.

\bibitem[Xiao et~al.(2022)Xiao, Huang, Guan, Zhan, and Lu]{xiao2022transfer}
Aoran Xiao, Jiaxing Huang, Dayan Guan, Fangneng Zhan, and Shijian Lu.
\newblock Transfer learning from synthetic to real lidar point cloud for
  semantic segmentation.
\newblock In \emph{Proceedings of the AAAI conference on artificial
  intelligence}, pages 2795--2803, 2022.

\bibitem[Xiong et~al.(2023)Xiong, Ma, Wang, and Urtasun]{xiong2023learning}
Yuwen Xiong, Wei-Chiu Ma, Jingkang Wang, and Raquel Urtasun.
\newblock Learning compact representations for lidar completion and generation.
\newblock In \emph{Proceedings of the IEEE/CVF Conference on Computer Vision
  and Pattern Recognition}, pages 1074--1083, 2023.

\bibitem[Yang et~al.(2019)Yang, Huang, Hao, Liu, Belongie, and
  Hariharan]{yang2019pointflow}
Guandao Yang, Xun Huang, Zekun Hao, Ming-Yu Liu, Serge Belongie, and Bharath
  Hariharan.
\newblock Pointflow: 3d point cloud generation with continuous normalizing
  flows.
\newblock In \emph{Proceedings of the IEEE/CVF international conference on
  computer vision}, pages 4541--4550, 2019.

\bibitem[Yuan et~al.(2023)Yuan, Cheng, Zeng, Su, Liu, Yu, and
  Wang]{yuan2023prototype}
Zhimin Yuan, Ming Cheng, Wankang Zeng, Yanfei Su, Weiquan Liu, Shangshu Yu, and
  Cheng Wang.
\newblock Prototype-guided multitask adversarial network for cross-domain lidar
  point clouds semantic segmentation.
\newblock \emph{IEEE Transactions on Geoscience and Remote Sensing},
  61:\penalty0 1--13, 2023.

\bibitem[Yuan et~al.(2024)Yuan, Zeng, Su, Liu, Cheng, Guo, and
  Wang]{yuan2024density}
Zhimin Yuan, Wankang Zeng, Yanfei Su, Weiquan Liu, Ming Cheng, Yulan Guo, and
  Cheng Wang.
\newblock Density-guided translator boosts synthetic-to-real unsupervised
  domain adaptive segmentation of 3d point clouds.
\newblock In \emph{Proceedings of the IEEE/CVF Conference on Computer Vision
  and Pattern Recognition}, pages 23303--23312, 2024.

\bibitem[Zamorski et~al.(2020)Zamorski, Zi{\k{e}}ba, Klukowski, Nowak, Kurach,
  Stokowiec, and Trzci{\'n}ski]{zamorski2020adversarial}
Maciej Zamorski, Maciej Zi{\k{e}}ba, Piotr Klukowski, Rafa{\l} Nowak, Karol
  Kurach, Wojciech Stokowiec, and Tomasz Trzci{\'n}ski.
\newblock Adversarial autoencoders for compact representations of 3d point
  clouds.
\newblock \emph{Computer Vision and Image Understanding}, 193:\penalty0 102921,
  2020.

\bibitem[Zhang et~al.(2023)Zhang, Rao, and Agrawala]{zhang2023adding}
Lvmin Zhang, Anyi Rao, and Maneesh Agrawala.
\newblock Adding conditional control to text-to-image diffusion models.
\newblock In \emph{Proceedings of the IEEE/CVF international conference on
  computer vision}, pages 3836--3847, 2023.

\bibitem[Zhao et~al.(2021)Zhao, Jiang, Jia, Torr, and Koltun]{zhao2021point}
Hengshuang Zhao, Li Jiang, Jiaya Jia, Philip~HS Torr, and Vladlen Koltun.
\newblock Point transformer.
\newblock In \emph{Proceedings of the IEEE/CVF international conference on
  computer vision}, pages 16259--16268, 2021.

\bibitem[Zhou et~al.(2021)Zhou, Du, and Wu]{zhou20213d}
Linqi Zhou, Yilun Du, and Jiajun Wu.
\newblock 3d shape generation and completion through point-voxel diffusion.
\newblock In \emph{Proceedings of the IEEE/CVF international conference on
  computer vision}, pages 5826--5835, 2021.

\bibitem[Zhu et~al.(2017)Zhu, Park, Isola, and Efros]{zhu2017unpaired}
Jun-Yan Zhu, Taesung Park, Phillip Isola, and Alexei~A Efros.
\newblock Unpaired image-to-image translation using cycle-consistent
  adversarial networks.
\newblock In \emph{Proceedings of the IEEE international conference on computer
  vision}, pages 2223--2232, 2017.

\bibitem[Zyrianov et~al.(2022)Zyrianov, Zhu, and Wang]{zyrianov2022learning}
Vlas Zyrianov, Xiyue Zhu, and Shenlong Wang.
\newblock Learning to generate realistic lidar point clouds.
\newblock In \emph{European Conference on Computer Vision}, pages 17--35.
  Springer, 2022.

\end{thebibliography}
}

\clearpage
\setcounter{page}{1}
\maketitlesupplementary
\section{Implementation} \label{sec: supp_id}
We build our model on a standard 2D diffusion framework \cite{ho2020denoising}, using a 2D U-Net \cite{ronneberger2015u} as the autoencoder backbone. Since lidar range images exhibit a wrap-around structure the same as panoramic images, we replace traditional convolutions with circular convolutions \cite{schubert2019circular}, following prior lidar diffusion models \cite{zyrianov2022learning,ran2024towards,hu2024rangeldm}. Additionally, we employ a lightweight three-layer CNN (the semantic projector) to map the U-Net’s latent space to the same resolution as the rescaled semantic map. For inference and lidar translation, we use DDIM \cite{song2021denoising}, a commonly adopted technique for efficient sampling. 

\section{Range Image and Point Cloud Conversion} \label{sec: supp_range}
Lidar range image leverages spherical projection to convert 3D point clouds into 2D images. Although there are some loss to this conversion, this technique has been shown effective to both the discriminative \cite{milioto2019rangenet++} and generative models \cite{caccia2019deep} for lidar data. Given each 3D point $(x,y,z)$ in the lidar coordinates, we have
\begin{itemize}
    \item Range:
    \begin{equation}
        r = \sqrt{x^2 + y^2 + z^2}
    \end{equation}
    \item Azimuth angle:
    \begin{equation}
        \theta = \operatorname{atan2}(y, x)
    \end{equation}
    \item Elevation angle:
    \begin{equation}
        \phi = \arcsin\!\biggl(\frac{z}{r}\biggr)
    \end{equation}
\end{itemize}
These angles are then rescaled and quantized to integer image 
coordinates $(u, v)$. For a $360^\circ$ sweep horizontally mapped into 
$u \in [0, 1023]$ and a set of 64 vertical rings mapped to 
$v \in [0, 63]$, we can apply
\begin{enumerate}
    \item \textbf{Horizontal index}
    \begin{equation}
        u = \left\lfloor \frac{1024}{2\pi}\, (\theta + \pi) \right\rfloor \in [0, 1023]
    \end{equation}
so that $\theta = -\pi$ goes to $u = 0$ and $\theta = +\pi$ goes near $u = 1023.$
\item \textbf{Vertical index}
\begin{equation}
    v = \left\lfloor \frac{64}{\phi_{\max} - \phi_{\min}} \, (\phi - \phi_{\min}) \right\rfloor 
\;\in [0, 63]
\end{equation}
where $\phi_{\min}, \phi_{\max}$ are the minimum/maximum elevation angles of the lidar
($-25^\circ$ to $+3^\circ$ for both SemanticKITTI and SynLiDAR).
\end{enumerate}
Finally, we can store the measured range $r$ (and possibly intensity and semantic labels) and  in the resulting 2D range image at pixel $(u, v)$.

\section{Evaluation Metrics} \label{sec:supp_em}
This section discusses the evaluation metrics used in the main body of the paper for assessing the quality of the generated point clouds in terms of both fidelity and diversity. The metrics for data generation can be categorized into two classes: perceptual and statistical.

\textbf{Perceptual metrics} measure the distance between real and generated data by comparing their representations in a perceptual space, which is derived from visual data using a pretrained feature extractor. In this research, we employ three perceptual metrics—FRID, FSVD, and FPVD—which serve as the lidar version of the commonly used Fréchet inception distance (FID).
\begin{itemize}
    \item \textbf{FRID} employs RangeNet++~\cite{milioto2019rangenet++}, a range-based lidar representation learning method, to extract features and compute distances. It is used as the primary metric because it evaluates only the regions within the range image, intentionally excluding areas outside where the data is less controlled. This approach reduces the influence of extraneous noise from regions far from the ego vehicle.
    \item \textbf{FSVD} employs MinkowskiNet \cite{choy20194d} to extract features by first voxelize the 3D point clouds. This method can cover the entire lidar space. The final feature vector is computed by averaging all non-empty voxel features from every point cloud segment.
    \item \textbf{FPVD} empoloys SPVCNN \cite{tang2020searching}, a point-voxel-based feature extractor which aggregates both point and volumetric features. This method can cover more geometric feature but in the other hand will be impacted more by the noisy points. The final feature vector is computed in the same way as FSVD.
\end{itemize}

\textbf{Statistical metrics} have been used as evaluation criteria for point cloud generative models since the pioneering work \cite{achlioptas2018learning}. These metrics rely on distance functions to quantify the similarity between pairs of point clouds. Among these, the Chamfer Distance (CD) has been the prevalent choice in recent studies~\cite{xiang2024synthetic,ran2024towards} due to its computational efficiency compared to other measures:
\begin{equation}\label{Eq:CD}
    CD(X,\hat{X})=\sum_{x\in X}\min_{y\in \hat{X}}\left \| x-y \right \|_{2}^{2}+\sum_{y\in \hat{X}}\min_{x\in X}\left \| x-y \right \|_{2}^{2}
\end{equation}
where $X$, $\hat{X}$ are the input and the reconstructed point cloud respectively, and $x, y$ are individual points. The Chamfer Distance is also used for evaluating the quality of the synthetic point clouds generated by the models. Based on this we have two metrics that focus on diversity and fidelity respectively:
\begin{itemize}
    \item \textbf{Jensen-Shannon Divergence (JSD)}  measures the similarity between two empirical distribution $P_{A}$ and $P_{B}$ based on the KL-divergence.\begin{equation}\label{Eq:JSD}
        \text{JSD}(P_{A}||P_{B})=\frac{1}{2}D_{KL}(P_{A}||M)+\frac{1}{2}D_{KL}(P_{B}||M)
    \end{equation}
    where $M=\frac{1}{2}(P_{A}+P_{B})$ and $D_{KL}(\cdot||\cdot)$ is the KL-divergence of distributions represented by two probability density functions, $P_{A}$ and $P_{B}$.
    \item \textbf{Minimum Matching Distance (MMD)} computes the average minimum distance between two matching point clouds from sets $S_{g}$ and $S_{r}$:
    \begin{equation}
        \text{MMD}(S_{g},S_{r})=\frac{1}{|S_{r}|}\sum_{Y\in S_{r}}\min_{X\in S_{g}}CD(X,Y)
    \end{equation}
\end{itemize}
Statistical metrics were originally designed for object-level point clouds, making them less suited to the more complex, scene-level data we work with. To address this mismatch, we follow the method in \cite{ran2024towards} by voxelizing the lidar point clouds and computing the metrics based on these voxels. Furthermore, metrics such as MMD are highly sensitive to noise. Since lidar data often includes uncontrollable noisy points, particularly in regions not captured by the range image. As a result, we place less emphasis on these statistical metrics compared to perceptual metrics.

\end{document}